\documentclass{article}

\usepackage{iclr2027_conference,times}
\usepackage[T1]{fontenc}
\usepackage{amsmath,amssymb,amsfonts,bm}
\usepackage{graphicx}
\usepackage{wrapfig}
\usepackage{booktabs}
\usepackage{multirow,bigstrut,threeparttable}
\usepackage{makecell,array}
\usepackage{colortbl,xcolor}
\usepackage{enumitem}
\usepackage{arydshln}
\usepackage{algorithm}
\usepackage{algorithmic}
\usepackage{adjustbox}
\usepackage{microtype}
\usepackage{textcomp}
\usepackage{verbatim}
\usepackage{url}
\usepackage[colorlinks=true,linkcolor=blue,citecolor=magenta,urlcolor=magenta]{hyperref}

\newcommand{\spm}{\tiny$\pm$~}

\usepackage{amsmath,amsfonts,bm}

\def\eqref#1{equation~\ref{#1}}

\def\1{\bm{1}}

\DeclareMathAlphabet{\mathsfit}{\encodingdefault}{\sfdefault}{m}{sl}
\SetMathAlphabet{\mathsfit}{bold}{\encodingdefault}{\sfdefault}{bx}{n}

\usepackage{amsthm}

\title{DeMMO: Cross-Disease Longitudinal \mbox{Modelling} of Digital Mobility Outcomes via Multi-Task Learning}

\author{Menghui Zhou \quad Zhipeng Yuan \quad Vitaveska Lanfranchi \quad Po Yang \\
School of Computer Science, University of Sheffield, Sheffield, United Kingdom \\
\texttt{\{menghui.zhou,zhipeng.yuan,v.lanfranchi,po.yang\}@sheffield.ac.uk}}

\iclrfinalcopy

\begin{document}

\maketitle
\lhead{}

\begin{abstract}

Digital mobility outcomes (DMOs) derived from wearable sensors characterise mobility in daily life and offer a promising means of monitoring disease progression.
However, existing DMO studies have typically focused on either a single disease or
a single visit.  To the best of our knowledge,
we are the first to define and study the practical problem of cross-disease longitudinal
DMO modelling. We argue that this problem should satisfy at least two requirements.
First, the temporal progression of DMOs should be modelled within each disease,
as mobility-limiting diseases evolve over time. Second, multiple
mobility-limiting diseases should be modelled jointly, as different diseases
affect different aspects of human mobility. To address this problem, we propose
DeMMO, an interpretable framework for longitudinal, multi-disease, and
multi-outcome learning. Its central technical contribution is an interpretable
cross-disease and cross-outcome relation-learning mechanism that infers signed
relations directly from learned longitudinal DMO--outcome mappings, thereby
enabling selective information sharing across cohorts without requiring paired
participants. We evaluate DeMMO on the recently released, large-scale,
multicentre Mobilise-D dataset, which presents a challenging modelling setting
involving longitudinal observations, multiple clinical outcomes, and four
participant-disjoint cohorts representing distinct mobility-limiting diseases.
Compared with eight strong structural longitudinal and
deep-regression baselines, DeMMO achieves the best overall predictive performance
and outperforms the baselines for most individual outcomes. Stability selection
further identifies reliable longitudinal DMO patterns that can inform subsequent
clinical validation and disease monitoring. The implementation code is available at \url{https://github.com/menghui-zhou/DeMMO}.
\end{abstract}

\section{Introduction}
\label{sec:intro}

Disease monitoring typically relies on periodic clinical assessments that may
require specialised resources, depend on evaluator judgement, or be affected
by recall and individual perceptions
\citep{delgado2023listening,deane2014priority,port2021people,prince2008comparison}.
Assessments performed weeks or months apart provide only intermittent snapshots
and may miss gradual, fluctuating, or transient changes
\citep{babrak2019traditional,coran2020advancing}. These limitations are
particularly consequential in mobility-limiting diseases such as Parkinson's
disease (PD), multiple sclerosis (MS), chronic obstructive pulmonary disease
(COPD), and proximal femoral fracture (PFF), which impose substantial personal
and socioeconomic burdens
\citep{yang2020current,bouleau2022socioeconomic,boers2025forecasting,williamson2017costs}.
Mobility reflects disease severity, progression, treatment response, and
recovery, yet a brief supervised test measures what an individual is
\emph{capable of doing} under standardised conditions rather than what they
\emph{actually do} in daily life \citep{rochester2020roadmap}. A six-minute
walk test, for example, samples only approximately \(0.06\%\) of a week and
cannot capture the diversity of everyday mobility.

Wearable sensing makes it possible to study this otherwise unobserved
free-living behaviour. Inertial sensors record movement over extended periods,
and validated analytical algorithms transform acceleration and
angular-velocity signals into digital mobility outcomes (DMOs)
\citep{goldsack2020verification,cerreta2020digital}. These outcomes describe
daily-life walking in terms of amount, pattern, pace, rhythm, and variability,
capturing both \emph{how much} and \emph{how} a person walks
\citep{rochester2020roadmap,kluge2021consensus}. Unlike a controlled assessment,
free-living monitoring samples walking bouts across days, environments, and
behavioural contexts and can therefore provide a more representative view of
everyday mobility \citep{alcock2025algorithms}. This ecological richness also
makes the modelling problem harder: free-living DMOs reflect not only clinical
capacity but also context, behaviour, day-to-day variation, and incomplete
follow-up. Clinically useful models must distinguish persistent disease-related
patterns from this natural heterogeneity and determine whether the relevant DMO
relationships remain stable or change over time.

Current evidence does not yet resolve this problem. Most clinical DMO studies
evaluate individual measures at a single visit using correlations, linear
models, or group comparisons
\citep{polhemus2021walking,mikolaizak2022connecting,megaritis2026construct,eckert2026construct}.
The emerging longitudinal literature is still disease-specific and commonly
based on small samples or a few prespecified measures: studies in PD, PFF, and
MS have shown that selected free-living measures can track change, but do not
model the evolving joint relationship between a comprehensive DMO profile and
multiple clinical outcomes
\citep{mirelman2024digital,engdal2024real,poleur2026stride}. Longitudinal DMO
modelling therefore remains limited even within individual diseases
\citep{rabano2025digital,kirk2025systematic}. To the best of our knowledge,
cross-disease longitudinal DMO modelling has not previously been defined or
studied, let alone in the more challenging free-living setting. This setting
must accommodate diseases that affect different aspects of mobility, clinical
outcomes measured on different scales, distinct participant cohorts, and the 
contextual heterogeneity of daily-life behaviour.

We therefore provide the first definition of cross-disease longitudinal DMO
modelling, which should satisfy at least two requirements. First, as
mobility-limiting diseases evolve over time, the temporal progression of DMOs
should be modelled within each disease. Second, as different mobility-limiting
diseases affect different aspects of human mobility, multiple diseases should be
modelled jointly to provide a more robust and comprehensive understanding of
mobility impairment.

A major barrier to studying this problem has been the lack of large, harmonised
longitudinal datasets that combine repeated measurements of a common DMO set
with multiple clinical outcomes across diseases
\citep{rochester2020roadmap}. Earlier resources were commonly cross-sectional,
disease-specific, heterogeneous in sensing and aggregation, or limited to few
assessment periods, preventing systematic analysis of stable, visit-specific,
and cross-disease DMO patterns.
\begin{wrapfigure}{r}{0.60\textwidth}
	\centering
	\vspace{-0.8\baselineskip}
	\includegraphics[width=\linewidth]{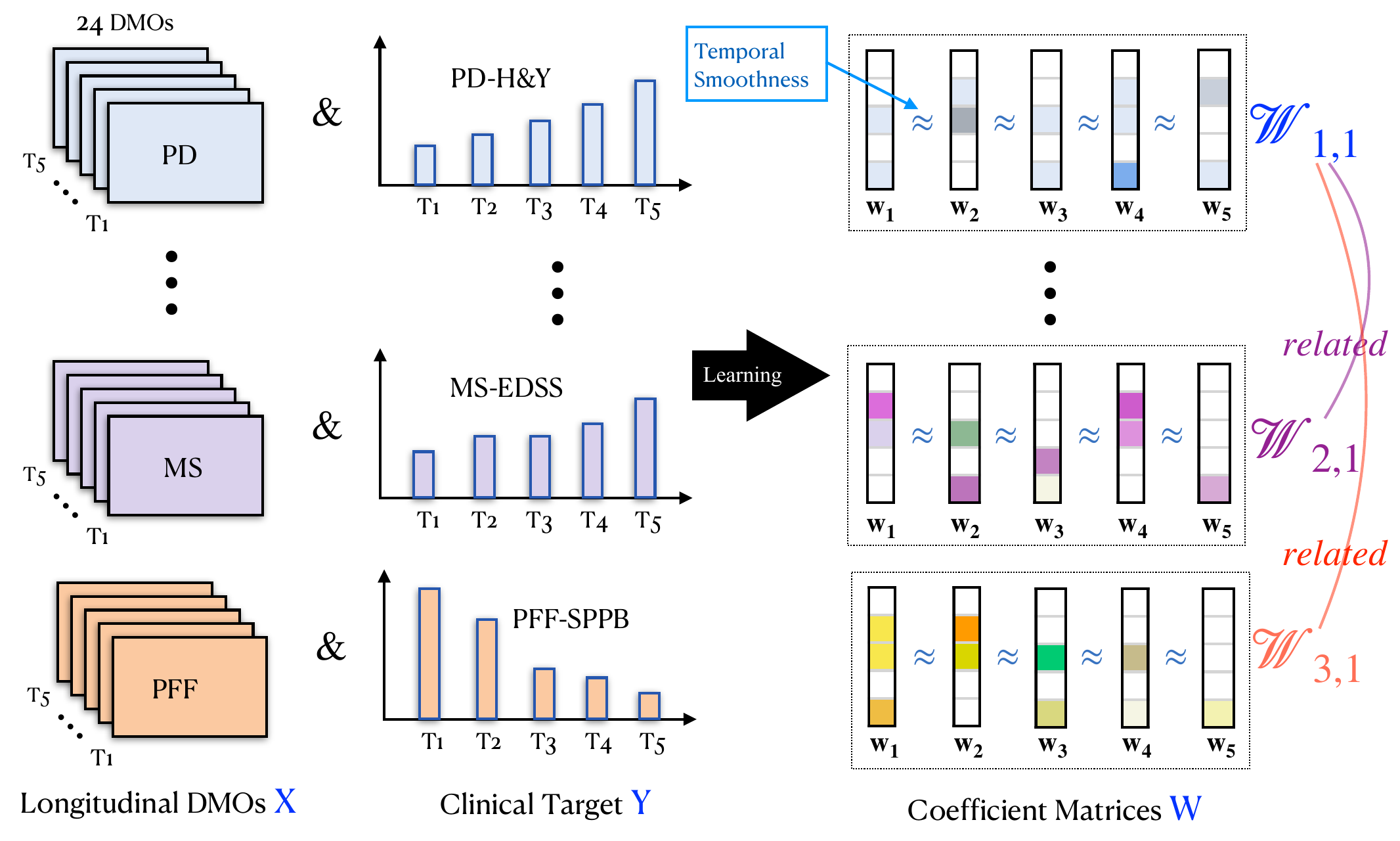}
	\caption{Overview of DeMMO. For each outcome, visit-specific coefficients map
		24 longitudinal DMOs to clinical targets. DeMMO jointly learns temporally
		smooth mappings and selective within- and cross-disease relations.}
	\label{fig:dummo}
\end{wrapfigure}
The recently released Mobilise-D dataset was
developed to address this gap
\citep{mikolaizak2022connecting,rochester2020roadmap, kirk2024mobilise}.
It is the principal data resource produced by a major European public--private
programme involving 35 partners from 13 countries. Conducted from 2019 to
2024, the programme represented a total investment of approximately
EUR~49.9 million. Mobilise-D first established the
technical validity of its sensor and analytical pipeline through a separate
multicentre study of 108 participants spanning healthy ageing and five
mobility-limiting conditions \citep{mikolaizak2022connecting}. The subsequent CVS
extended this foundation to clinical validation at an unprecedented large scale:
2,400 participants across PD, MS, COPD, and PFF underwent a comprehensive
clinical assessment and seven days of real-world mobility monitoring on five
occasions, separated by six-month intervals over two years \citep{kirk2024mobilise}. The released resource combines harmonised
walking-bout, daily, and weekly DMOs from a common lower-back sensor with
clinical tests, clinician-reported assessments, and patient-reported outcomes.
This combination of scale, longitudinal depth, cross-disease breadth, and
standardised measurement makes Mobilise-D uniquely suited to studying how
multivariate DMO--outcome relationships evolve during disease progression and
functional recovery.

%
%

We summarise the main contributions of this work as follows:

\begin{itemize}
\item \textbf{Conceptual contribution.} To the best of our knowledge, we are
the first to define cross-disease longitudinal DMO modelling. The problem
establishes two fundamental requirements: modelling temporal DMO progression
within each disease and jointly modelling multiple diseases to capture their
complementary effects on human mobility. We further present the first
systematic study of this problem in the more challenging free-living setting.

\item \textbf{Technical contribution.} We expose a key limitation of existing
temporal multi-task learning: current formulations do not jointly learn
multiple outcomes across diseases with non-overlapping cohorts. To address
this limitation, we introduce DeMMO and a novel cross-disease and cross-outcome
relation-learning mechanism. As shown in Figure~\ref{fig:dummo}, DeMMO learns a
signed relation graph directly from longitudinal DMO mappings, enabling
selective knowledge transfer across unpaired cohorts while retaining
outcome-specific temporal evolution and feature selection.

\item \textbf{Practical contribution.} Using the recently released Mobilise-D
dataset, we jointly study multiple clinical outcomes over five visits across
several mobility-limiting diseases. Against eight strong baselines spanning
structured longitudinal and recent deep-regression methods, DeMMO achieves the
best overall predictive performance and outperforms the baselines for most
individual outcomes. Longitudinal stability selection
\citep{zhou2024integrating,fan2026beyond} further identifies robust,
outcome-specific DMO patterns as candidates for subsequent clinical validation
and disease monitoring.
\end{itemize}

Taken together, to the best of our knowledge, this work provides the first
definition and systematic study of cross-disease longitudinal DMO modelling,
let alone its more challenging free-living form.
We  emphasise that the simplicity, concreteness, and high
interpretability of DeMMO make the framework readily understandable beyond the
computer science community, thereby facilitating its broader acceptance and
adoption by both computational researchers and non-computational domain experts,
particularly clinicians.
 Further discussion of related work and DeMMO's potential use in interpretable mobility-monitoring software is provided in Appendices~\ref{sec:related} and~\ref{sec:clinical_application}, respectively.

\section{Method}
\label{sec:method}

\subsection{Mobilise-D Longitudinal Dataset}
\label{sec:mobilise_dataset}
To ground the proposed methodology in a concrete setting, we first introduce
the recently released Mobilise-D dataset, a multicentre
longitudinal DMO study of people with mobility-limiting conditions
\citep{mazza2021technical,mikolaizak2022connecting}. The dataset contains
mutually exclusive PD, MS, COPD, and PFF cohorts. Participants were scheduled
for five visits, \(T_1,\ldots,T_5\), followed by real-world monitoring with a
body-worn inertial sensor. 
\begin{wraptable}{r}{0.65\textwidth}
	\centering
	\caption{Mobilise-D analytical dataset after requiring a valid target and complete observations for all 24 weekly DMOs. $N$ denotes participants with at least one retained visit.}
	\label{tab:dataset_statistics}
	\resizebox{\linewidth}{!}{
		\begin{tabular}{llrrrrrrrr}
			\toprule
			Disease & Target & $N$ & \(T_1\) & \(T_2\) & \(T_3\) & \(T_4\) & \(T_5\) & Total & Range \\
			\midrule
			PD  & H\&Y stage          & 574 & 528 & 439 & 400 & 363 & 346 & 2,076 & 0--4 \\
			PD  & MDS--UPDRS III      & 574 & 528 & 439 & 400 & 363 & 346 & 2,076 & 1--84 \\
			MS  & EDSS                & 578 & 535 & 448 & 402 & 346 & 352 & 2,083 & 0--7.5 \\
			PFF & SPPB impairment     & 469 & 418 & 358 & 282 & 146 & 108 & 1,312 & 0--11 \\
			COPD & FEV$_1$ \% predicted & 583 & 547 & 437 & 376 & 364 & 348 & 2,072 & 9.26--117.18 \\
			\bottomrule
		\end{tabular}
	}
\end{wraptable}
We analyse all four cohorts at all five visits. We
use filtered weekly aggregates containing at least three valid monitoring
days. Each record contains 24 DMOs describing walking amount, bout pattern,
pace, rhythm, and variability. The quality-control variable
\texttt{n\_days\_w} is not used as a predictor. Definitions of all 24 DMOs
are provided in Appendix~\ref{sec:dmo_definitions}.
We consider five regression objectives: H\&Y stage (PD H\&Y)
\citep{hoehn1967parkinsonism} and MDS--UPDRS Part~III (PD MDS--UPDRS)
\citep{goetz2008movement} for PD, EDSS for MS (MS EDSS)
\citep{kurtzke1983rating}, SPPB for PFF (PFF SPPB)
\citep{bellettiere2020short}, and FEV$_1$ as a percentage of its predicted
value for COPD (COPD FEV$_1$) \citep{agusti2023global}.  Because higher SPPB indicates better
function, we define SPPB impairment as \(12-\mathrm{SPPB}\), while retaining
PFF SPPB as the objective's short name. Invalid values, including
MDS--UPDRS scores of \(-1\), are treated as missing.
Complete-case processing is applied at the participant--visit level. A record is retained only when its clinical outcome and all 24 DMOs are available. Table~\ref{tab:dataset_statistics} summarises the resulting dataset. Sample sizes decrease across visits because of loss to follow-up, unavailable recordings, insufficient monitoring days, and incomplete clinical assessments.

\subsection{Longitudinal Clinical-Progression Formulation}
\label{sec:progression_formulation}

We first consider one clinical outcome from a single disease cohort. This simplified setting introduces the longitudinal modelling, temporal regularisation, and DMO-selection components before their extension to multiple diseases and outcomes in Section~\ref{sec:multi_disease_multi_outcome}.

At visit \(t\), let \(\mathbf{X}_t\in\mathbb{R}^{n_t\times p}\) denote the DMO feature matrix for \(n_t\) retained records, where \(p=24\), and let \(\mathbf{y}_t\in\mathbb{R}^{n_t}\) contain the corresponding clinical outcomes. Unlike conventional progression models that use baseline measurements to predict future outcomes, Mobilise-D provides both DMOs and clinical outcomes at each of \(T=5\) visits. We therefore treat clinical-status estimation at each visit as a separate but related regression task:
$
	\mathbf{y}_t=\mathbf{X}_t\mathbf{w}_t+\boldsymbol{\epsilon}_t,
$
where \(\mathbf{w}_t\in\mathbb{R}^{p}\) is the visit-specific coefficient vector and \(\boldsymbol{\epsilon}_t\) denotes observational noise. The coefficients are collected as \(\mathbf{W}=[\mathbf{w}_1,\ldots,\mathbf{w}_T]\in\mathbb{R}^{p\times T}\). Each column represents the DMO--outcome relationship at one visit, while each row describes the longitudinal coefficient trajectory of one DMO.
We use squared-error regression because the clinical outcomes are numerical scores. To prevent visits with larger samples from dominating estimation, the loss is normalised within each visit:
$
	\mathcal{L}(\mathbf{W})
	=
	\sum_{t=1}^{T}
	\frac{1}{2n_t}
	\left\|
	\mathbf{y}_t-\mathbf{X}_t\mathbf{w}_t
	\right\|_2^2.
$

\paragraph{Temporal Smoothness.}
We model temporal smoothness using the fused Lasso, a well-established
regulariser for longitudinal biomarker analysis
\citep{zhou2012modeling,zhou2023robust}. Let
$\mathbf{R}\in\mathbb{R}^{(T-1)\times T}$ be the first-order difference
matrix, with $R_{i,i}=-1$, $R_{i,i+1}=1$, and all other entries zero. We use
\(\Omega_{\mathrm{FL}}(\mathbf{W})=\|\mathbf{W}\mathbf{R}^{\mathsf T}\|_1\),
which encourages adjacent visits to share coefficients while retaining
temporally localised changes. This convex structure and its efficient
optimisation have been established in previous work
\citep{zhou2012modeling,zhou2023robust}.

\paragraph{Sparse Group Lasso.}

Following \citet{zhou2011multi}, we use sparse group Lasso for interpretable
DMO selection. Its element-wise
term selects visit-specific effects, whereas its row-wise term selects DMOs
across the longitudinal period:
$
	\Omega_{\mathrm{SGL}}(\mathbf{W})
	=
	\lambda_{\mathrm{L}}\|\mathbf{W}\|_1
	+
	\lambda_{\mathrm{G}}\|\mathbf{W}\|_{2,1},
$
where \(\|\mathbf{W}\|_{2,1}=\sum_{j=1}^{p}\|\mathbf{w}_{j:}\|_2\).
Thus, \(\lambda_{\mathrm{L}}\) controls visit-specific sparsity and
\(\lambda_{\mathrm{G}}\) controls DMO-level sparsity. Sparse group Lasso is a
mature convex feature-selection structure with efficient proximal algorithms
\citep{tibshirani1996regression,zhou2022multi}; selected DMOs are interpreted
as candidates requiring subsequent clinical validation.

Combining the regression loss, fused temporal regularisation, and sparse group
Lasso yields cFSGL, a classical temporal multi-task learning formulation
\citep{zhou2012modeling, zhou2013modeling}:
\begin{equation}
	\min_{\mathbf{W}}\quad
	\sum_{t=1}^{T}\frac{1}{2n_t}\|\mathbf{y}_t-\mathbf{X}_t\mathbf{w}_t\|_2^2
	+\lambda_{\mathrm{FL}}\|\mathbf{W}\mathbf{R}^{\mathsf T}\|_1
	+\lambda_{\mathrm{L}}\|\mathbf{W}\|_1
	+\lambda_{\mathrm{G}}\|\mathbf{W}\|_{2,1}.
	\label{eq:single-outcome}
\end{equation}
cFSGL has subsequently inspired numerous extensions of temporal multi-task
learning \citep{cao2017sparse,zhou2022multi,zhou2023automatic,zhou2023robust,fan2026beyond}.
However, this line of work models only a single clinical outcome within a single disease cohort.  It cannot jointly model
multiple outcomes across diseases, let alone accommodate disease cohorts with
non-overlapping participants. We next extend this formulation to address both
settings.

\subsection{Multi-Disease and Multi-Outcome Extension}
\label{sec:multi_disease_multi_outcome}

Our analysis includes multiple prediction objectives across four cohorts: PD
contributes two clinical outcomes, whereas MS, PFF, and COPD each contribute
one. The number of prediction objectives therefore differs from the number of
diseases. Let \(d\in\{1,\ldots,D\}\) index the disease cohorts, where \(D=4\),
and let \(Q_d\) denote the number of clinical outcomes for disease \(d\). The
total number of disease--outcome prediction objectives is
\(M=\sum_{d=1}^{D}Q_d=5\).

For disease \(d\), outcome \(q\), and visit \(t\), let \(\mathbf{X}_{t,q}^{(d)}\in\mathbb{R}^{n_{dtq}\times p}\) contain the complete DMO observations and let \(\mathbf{y}_{t,q}^{(d)}\in\mathbb{R}^{n_{dtq}}\) contain the corresponding clinical outcomes. Here, \(n_{dtq}\) is the number of retained participant--visit records. Although all objectives share the same 24 DMO definitions, allowing \(\mathbf{X}_{t,q}^{(d)}\) to depend on \(q\) accommodates outcome-specific differences in data availability.
The visit-specific regression model is
\begin{equation}
	\mathbf{y}_{t,q}^{(d)}
	=
	\mathbf{X}_{t,q}^{(d)}
	\mathbf{w}_{t,q}^{(d)}
	+
	\boldsymbol{\epsilon}_{t,q}^{(d)},
\end{equation}
where \(\mathbf{w}_{t,q}^{(d)}\in\mathbb{R}^{p}\). The longitudinal coefficient matrix for disease \(d\) and outcome \(q\) is \(\mathbf{W}_{d,q}=[\mathbf{w}_{1,q}^{(d)},\ldots,\mathbf{w}_{T,q}^{(d)}]\in\mathbb{R}^{p\times T}\). Thus, each \(\mathbf{W}_{d,q}\) represents one longitudinal clinical prediction objective rather than an entire disease.

\subsection{Automatic Relation Learning Across Prediction Objectives}
\label{sec:automatic_relation_learning}

PD, MS, PFF, and COPD have distinct pathological mechanisms and clinical
outcomes, but their DMO--outcome mappings may share functional patterns.
Similarly, the two PD outcomes may reflect related but non-identical aspects
of disease severity and motor impairment.
These relationships cannot be specified reliably in advance. Outcomes from different diseases are not jointly observed because the cohorts contain different participants, while correlations between outcomes within the same disease do not necessarily reflect similarities in their longitudinal DMO mappings. We therefore learn relationships among prediction objectives directly from their model parameters.

For disease \(d\) and outcome \(q\), let \(\mathbf{u}_{d,q}=\operatorname{vec}(\mathbf{W}_{d,q})\in\mathbb{R}^{pT}\). The representations of all \(M\) prediction objectives are collected as \(\mathbf{U}\in\mathbb{R}^{pT\times M}\), with each column corresponding to one disease--outcome pair.
We assume that each longitudinal DMO mapping is related to the remaining
mappings. After assigning each disease--outcome pair a unique objective index,
this relation is expressed as
\begin{equation}
	\mathbf{U}\approx\mathbf{U}\mathbf{A},
	\qquad
	\mathbf{A}=\mathbf{A}^{\mathsf T},
	\qquad
	\operatorname{diag}(\mathbf{A})=\mathbf{0},
\end{equation}
where \(\mathbf{A}\in\mathbb{R}^{M\times M}\) is the learned relation matrix. Symmetry produces an undirected relation graph, while the zero diagonal excludes trivial self-relations.

The matrix \(\mathbf{A}\) can be partitioned into blocks \(\mathbf{A}^{(d,k)}\in\mathbb{R}^{Q_d\times Q_k}\) according to disease membership. A diagonal block \(\mathbf{A}^{(d,d)}\) captures relationships among outcomes within disease \(d\), whereas an off-diagonal block \(\mathbf{A}^{(d,k)}\) captures cross-disease relationships. Symmetry implies \(\mathbf{A}^{(d,k)}=(\mathbf{A}^{(k,d)})^{\mathsf T}\).
We estimate the relation structure using
\begin{equation}
	\Omega_{\mathrm{R}}(\mathbf{U},\mathbf{A})
	=
	\frac{\lambda_{\mathrm{R}}}{2}
	\left\|
	\mathbf{U}-\mathbf{U}\mathbf{A}
	\right\|_F^2
	+
	\frac{\lambda_{\mathrm{A}}}{2}
	\left\|
	\mathbf{A}
	\right\|_F^2,
\end{equation}
where \(\lambda_{\mathrm{R}}\geq0\) controls cross-objective information sharing and \(\lambda_{\mathrm{A}}>0\) stabilises relation estimation. We use an \(\ell_2\) penalty rather than sparsity because the model contains only five prediction objectives and weak relationships may still provide useful information.

\subsection{Unified Multi-Disease and Multi-Outcome Objective}
\label{sec:unified_objective}

Combining longitudinal regression, fused temporal regularisation, sparse group DMO selection, and cross-disease and
cross-outcome relation learning yields the proposed \emph{DMO-enabled Multi-Disease and Multi-Outcome learning} framework (DeMMO):

\begin{equation}
	\label{eq:dummo_objective}
	\begin{aligned}
		\min_{\{\mathbf{W}_{d,q}\},\,\mathbf{A}}\quad
		&\sum_{d=1}^{D}\sum_{q=1}^{Q_d}\sum_{t=1}^{T}
		\frac{\|\mathbf{y}_{t,q}^{(d)}-\mathbf{X}_{t,q}^{(d)}\mathbf{w}_{t,q}^{(d)}\|_2^2}{2n_{dtq}} \\
		&+\sum_{d=1}^{D}\sum_{q=1}^{Q_d}\!\left(
		\lambda_{\mathrm{FL}}\|\mathbf{W}_{d,q}\mathbf{R}^{\mathsf T}\|_1
		+\lambda_{\mathrm{L}}\|\mathbf{W}_{d,q}\|_1
		+\lambda_{\mathrm{G}}\|\mathbf{W}_{d,q}\|_{2,1}\right) \\
		&+\frac{\lambda_{\mathrm{R}}}{2}\|\mathbf{U}-\mathbf{U}\mathbf{A}\|_F^2
		+\frac{\lambda_{\mathrm{A}}}{2}\|\mathbf{A}\|_F^2 \\
		\text{\rm s.t.}\quad
		&\mathbf{A}=\mathbf{A}^{\mathsf T},\qquad
		\operatorname{diag}(\mathbf{A})=\mathbf{0}.
	\end{aligned}
\end{equation}
%

Notably, DeMMO does not require shared participants across disease cohorts, and each prediction objective retains its own longitudinal coefficient matrix. The objective is biconvex in \(\{\mathbf{W}_{d,q}\}\) and \(\mathbf{A}\). For fixed \(\mathbf{A}\), the coefficient subproblem is convex but non-smooth because of the fused-Lasso, Lasso, and group-Lasso penalties. For fixed coefficient matrices, relation learning is a strongly convex quadratic problem over symmetric matrices with zero diagonal. We alternate between these two subproblems.

\subsection{Data Standardisation}
\label{sec:data_standardisation}

PD H\&Y, PD MDS--UPDRS, MS EDSS, and COPD FEV$_1$ retain
their original directions, while SPPB is transformed as
\(12-\mathrm{SPPB}\) to represent impairment. Consequently, higher values
indicate greater severity or impairment for the first four objectives, whereas
higher COPD FEV$_1$ indicates better pulmonary function. Each outcome
is then standardised using the mean and standard deviation estimated from its
training records pooled across visits.

The 24 DMOs are similarly standardised using pooled training records from the corresponding disease cohort. Pooling across visits preserves longitudinal distributional changes that visit-specific standardisation could remove. All transformations and standardisation parameters are estimated exclusively from the training data and applied unchanged to the validation and test sets to prevent information leakage.

\paragraph{Additional results in the appendix.}
Due to space constraints, the complete alternating optimisation procedure,
convergence analysis, and computational complexity are deferred to
Appendix~\ref{sec:optimization}. The resulting algorithm is computationally
efficient:
coefficient updates use established accelerated proximal-gradient operations,
while relation learning adds little overhead because the five outcomes yield
only ten free relation parameters. Because convex sparsity-inducing penalties
are known to shrink nonzero coefficients and may introduce estimation bias
\citep{fan2001variable}, Appendix~\ref{sec:noncon} develops two non-convex
variants, DeMMO-var1 and DeMMO-var2, designed to reduce this bias. Their empirical results are reported in
Appendix~\ref{sec:nonconvex_variant_results}. Despite their greater
computational cost, both non-convex variants perform worse overall than the
original DeMMO, suggesting that reducing sparsity-induced estimation bias is
not beneficial in this setting. We conjecture that, because free-living DMOs
contain substantial behavioural and measurement noise, the stronger shrinkage
of the convex formulation provides useful regularisation, whereas the
non-convex variants may retain more noise and generalise less effectively.

\section{Experiments}
\label{sec:exp}

We evaluate DeMMO on the Mobilise-D dataset described in
Section~\ref{sec:mobilise_dataset}. The analysis includes 24 weekly DMOs
measured over five visits and five clinical prediction objectives: PD H\&Y,
PD MDS--UPDRS, MS EDSS, PFF SPPB, and COPD FEV$_1$. The resulting
objective-specific sample sizes are reported in
Table~\ref{tab:dataset_statistics}.
For each experimental run, participants are divided into training,
validation, and held-out test sets in a 70\%/10\%/20\% ratio.
All available visits from the same participant remain in one split to prevent
information leakage. Data processing and standardisation follow
Section~\ref{sec:data_standardisation}, with all parameters estimated from the
training data only. We repeat the complete procedure over five matched
participant-level splits using random seeds 42--46.
Following common practice \citep{zhou2024integrating,fan2026beyond}, we use 
normalised mean squared error (nMSE) and weighted correlation (wR) to assess
overall performance, and root mean squared error (RMSE) for visit-specific
performance.

\subsection{Compared Methods and Training Settings}
\label{sec:baselines}

We compare DeMMO with eight baselines spanning conventional
interpretable longitudinal models, and recent deep regression methods. All
methods follow the common data-splitting and preprocessing protocol described
above, with hyperparameters selected on the validation set and the held-out
test set reserved for final evaluation. Methods that do not explicitly model
time are fitted independently to each of the 25 outcome--visit tasks.

\paragraph{DeMMO.}
DeMMO is trained using the alternating optimisation procedure described in
Section~\ref{sec:optimization}. Because a joint grid search over its five
regularisation parameters is computationally expensive, we use two-stage
selection. First, relation learning is disabled while
$\lambda_{\mathrm{FL}}$, $\lambda_{\mathrm{L}}$, and
$\lambda_{\mathrm{G}}$ are selected from
$\{10^{-3},10^{-2},10^{-1},1\}$. These parameters are then fixed, and
$(\lambda_{\mathrm{R}},\lambda_{\mathrm{A}})$ is selected from $(0,0)$ and
$\{10^{-3},10^{-2},10^{-1}\}^2$. The selected model is refitted on the
combined training and validation sets. Optimisation is limited to 50 outer and
500 inner iterations, with convergence tolerances of $10^{-6}$ and $10^{-7}$,
respectively.

\paragraph{Baselines.}
For cFSGL \citep{zhou2012modeling,zhou2013learning}, FRoTS
\citep{zhou2023robust}, and MAGPP \citep{zhou2024integrating}, all penalty
hyperparameters are selected from
$\{10^{-4},10^{-3},10^{-2},10^{-1}\}$ using validation nMSE.
MLP-MSE uses a two-hidden-layer MLP with 32 ReLU units per layer, MSE loss,
Adam optimisation, and validation-based early stopping. MLP-L1 uses the same
setting but replaces MSE with MAE. Rank-N-Contrast (Rank-N)
\citep{zha2024rank} pretrains a contrastive encoder using unit-width target
bins and then fits a linear regression head. RankSim \citep{gong2022ranksim}
combines MSE with a regulariser that aligns representation and continuous-label
rankings. ACCon \citep{zhao2025accon} combines MSE with an angle-compensated
contrastive regulariser, whose weight is selected from
$\{10^{-2},10^{-1},1,10,100\}$ using validation MSE.

\subsection{Hyperparameter Sensitivity}
\label{sec:hyperparameter_sensitivity}

\begin{figure*}[!t]
	\centering
	\includegraphics[width=0.98\textwidth]{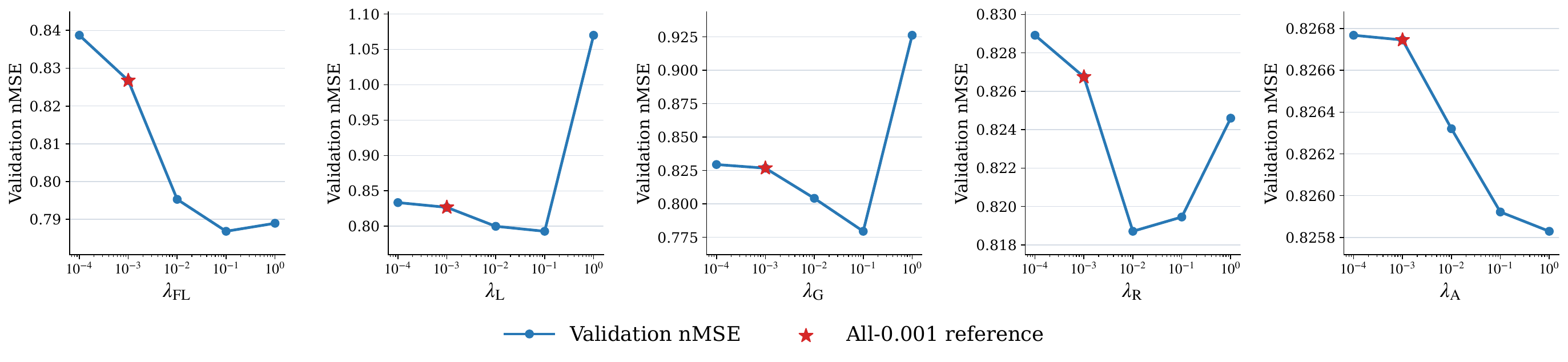}
	\caption{One-at-a-time sensitivity of validation nMSE; all other parameters
		are fixed at 0.001, with the shared reference marked by a red star.}
	\label{fig:hyperparameter_sensitivity}
\end{figure*}

We vary each parameter over $\{10^{-4},10^{-3},10^{-2},10^{-1},1\}$ while
fixing the others at $10^{-3}$, and report mean validation nMSE across the 25
outcome--visit tasks for seed 42 (Fig.~\ref{fig:hyperparameter_sensitivity});
the test set is not used. All five penalties affect validation performance.
The sparsity penalties are most influential and become excessive at 1,
$\lambda_{\mathrm{FL}}$ favours moderate temporal smoothing, and the U-shaped
$\lambda_{\mathrm{R}}$ curve supports selective rather than unrestricted
cross-objective sharing. Performance is comparatively insensitive to
$\lambda_{\mathrm{A}}$ over the tested range, likely because standardising the
predictors and outcomes places the coefficient matrices across disease--outcome
objectives on comparable scales.

\subsection{Prediction Performance and Learned Cross-Objective Relations}
\label{sec:prediction_performance}

\begin{table*}[!t]
	\centering
	\caption{Summary test performance on Mobilise-D (mean $\pm$ standard deviation). The best mean in each row is shown in bold.}
	\label{tab:mobilise_results_summary}
	\resizebox{\textwidth}{!}{%
		\begin{tabular}{llccccccccc}
			\toprule
			Objective & Metric & cFSGL & FRoTS & MAGPP & MLP-MSE & MLP-L1 & Rank-N & RankSim & ACCon & DeMMO \\
			\midrule
			\multirow{2}{*}{Overall}
			& nMSE $\downarrow$ & 0.787 \spm0.020 & 0.795 \spm0.020 & 0.812 \spm0.021 & 0.811 \spm0.023 & 0.835 \spm0.015 & 0.836 \spm0.019 & 0.807 \spm0.017 & 0.828 \spm0.020 & \cellcolor{blue!10}{\textbf{0.769 \spm0.021}} \\
			& wR $\uparrow$ & 0.443 \spm0.019 & 0.448 \spm0.015 & 0.437 \spm0.016 & 0.417 \spm0.024 & 0.399 \spm0.014 & 0.372 \spm0.023 & 0.421 \spm0.019 & 0.405 \spm0.017 & \cellcolor{blue!10}{\textbf{0.464 \spm0.023}} \\
			\midrule
			\multirow{2}{*}{PD H\&Y}
			& nMSE $\downarrow$ & 0.969 \spm0.021 & 0.970 \spm0.026 & 0.994 \spm0.029 & 0.990 \spm0.020 & 1.025 \spm0.013 & 1.035 \spm0.008 & 0.992 \spm0.021 & 1.008 \spm0.017 & \cellcolor{blue!10}{\textbf{0.937 \spm0.020}} \\
			& wR $\uparrow$ & 0.264 \spm0.025 & 0.275 \spm0.022 & 0.254 \spm0.023 & 0.214 \spm0.060 & 0.156 \spm0.021 & 0.095 \spm0.045 & 0.234 \spm0.042 & 0.234 \spm0.026 & \cellcolor{blue!10}{\textbf{0.300 \spm0.033}} \\
			\midrule
			\multirow{2}{*}{PD MDS--UPDRS}
			& nMSE $\downarrow$ & 0.899 \spm0.034 & 0.899 \spm0.035 & 0.917 \spm0.035 & 0.916 \spm0.030 & 0.956 \spm0.022 & 0.956 \spm0.029 & 0.928 \spm0.030 & 0.951 \spm0.036 & \cellcolor{blue!10}{\textbf{0.885 \spm0.032}} \\
			& wR $\uparrow$ & 0.347 \spm0.032 & 0.363 \spm0.027 & 0.351 \spm0.027 & 0.304 \spm0.034 & 0.285 \spm0.030 & 0.240 \spm0.052 & 0.298 \spm0.036 & 0.264 \spm0.032 & \cellcolor{blue!10}{\textbf{0.369 \spm0.034}} \\
			\midrule
			\multirow{2}{*}{MS EDSS}
			& nMSE $\downarrow$ & 0.497 \spm0.037 & 0.501 \spm0.037 & 0.514 \spm0.038 & 0.515 \spm0.034 & 0.527 \spm0.034 & 0.538 \spm0.036 & 0.508 \spm0.036 & 0.525 \spm0.036 & \cellcolor{blue!10}{\textbf{0.482 \spm0.034}} \\
			& wR $\uparrow$ & 0.715 \spm0.026 & 0.713 \spm0.025 & 0.706 \spm0.026 & 0.704 \spm0.023 & 0.703 \spm0.023 & 0.697 \spm0.027 & 0.708 \spm0.024 & 0.697 \spm0.023 & \cellcolor{blue!10}{\textbf{0.731 \spm0.024}} \\
			\midrule
			\multirow{2}{*}{PFF SPPB}
			& nMSE $\downarrow$ & 0.560 \spm0.039 & 0.571 \spm0.052 & 0.585 \spm0.055 & 0.585 \spm0.036 & 0.592 \spm0.036 & 0.593 \spm0.043 & 0.579 \spm0.039 & 0.594 \spm0.055 & \cellcolor{blue!10}{\textbf{0.550 \spm0.040}} \\
			& wR $\uparrow$ & 0.671 \spm0.027 & 0.668 \spm0.034 & 0.662 \spm0.035 & 0.661 \spm0.023 & 0.661 \spm0.019 & 0.655 \spm0.027 & 0.660 \spm0.026 & 0.650 \spm0.040 & \cellcolor{blue!10}{\textbf{0.680 \spm0.029}} \\
			\midrule
			\multirow{2}{*}{COPD FEV$_1$}
			& nMSE $\downarrow$ & 0.933 \spm0.021 & 0.955 \spm0.020 & 0.969 \spm0.023 & 0.972 \spm0.054 & 0.989 \spm0.033 & 0.977 \spm0.031 & 0.951 \spm0.036 & 0.982 \spm0.044 & \cellcolor{blue!10}{\textbf{0.916 \spm0.017}} \\
			& wR $\uparrow$ & 0.298 \spm0.027 & 0.296 \spm0.018 & 0.289 \spm0.019 & 0.285 \spm0.035 & 0.278 \spm0.029 & 0.267 \spm0.032 & 0.286 \spm0.037 & 0.265 \spm0.036 & \cellcolor{blue!10}{\textbf{0.315 \spm0.027}} \\
			\bottomrule
		\end{tabular}%
	}
\end{table*}

\begin{figure*}[!t]
	\centering
	\includegraphics[width=\textwidth]{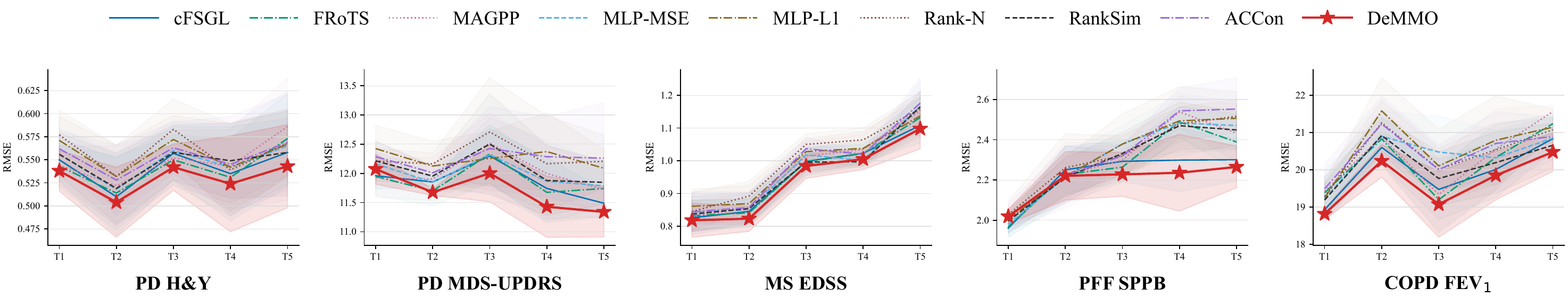}
	\caption{Visit-specific RMSE (mean $\pm$ standard deviation over five matched participant splits) for the five prediction objectives. Each panel reports T1--T5 performance for all baselines and DeMMO. Lower values indicate better predictive accuracy; no visit-level significance test is displayed.}
	\label{fig:rmse_across_visits}
\end{figure*}

\paragraph{Overall and outcome-specific performance.}
DeMMO achieves the best mean nMSE and wR both overall and for each of the five
prediction outcomes (Table~\ref{tab:mobilise_results_summary}). Relative to the
strongest baseline for each metric, DeMMO reduces overall nMSE from $0.787$ to
$0.769$ and increases overall wR from $0.448$ to $0.464$. Its advantage is
consistent across outcomes: the nMSE improvements over the best corresponding
baselines are $0.032$, $0.014$, $0.015$, $0.010$, and $0.017$ for PD H\&Y,
PD MDS--UPDRS, MS EDSS, PFF SPPB, and COPD FEV$_1$, respectively. The
corresponding wR improvements range from $0.006$ to $0.025$.
The comparison with cFSGL most directly isolates the value of cross-objective
relation learning. cFSGL uses the same temporal smoothing and stable and
visit-specific feature-selection structure, but fits each clinical objective
separately. By additionally learning and exploiting relations across
objectives, DeMMO improves overall nMSE from $0.787$ to $0.769$ and wR from
$0.443$ to $0.464$, while outperforming cFSGL on both metrics for all five
objectives. This consistent gain indicates that DeMMO's advantage arises from
selective cross-objective information sharing rather than temporal
regularisation alone. More broadly, the structured longitudinal methods
generally outperform the deep-regression baselines, suggesting that explicitly
modelling the problem structure is more valuable here than increasing model
capacity.

\paragraph{Visit-specific performance.}
Figure~\ref{fig:rmse_across_visits} shows that the aggregate improvement is not
driven by a single visit or outcome. DeMMO obtains the lowest mean RMSE in 21
of the 25 outcome--visit tasks: all five visits for PD H\&Y and COPD FEV$_1$,
T2--T5 for PD MDS--UPDRS, T1--T3 and T5 for MS EDSS, and T3--T5 for PFF SPPB.
Other methods lead only at PD MDS--UPDRS T1, MS EDSS T4, and PFF SPPB T1--T2.
The later-visit gains for PFF are particularly relevant because its follow-up
sample size decreases substantially. More broadly, the heterogeneous error
trajectories across outcomes show the need to preserve outcome-specific
behaviour while sharing information selectively. Detailed values are provided
in Appendix~\ref{sec:additional_results},
Tables~\ref{tab:visit_rmse_pd_hy}--\ref{tab:visit_rmse_copd}.

\begin{wrapfigure}{r}{0.54\textwidth}
	\centering
	\includegraphics[width=\linewidth]
	{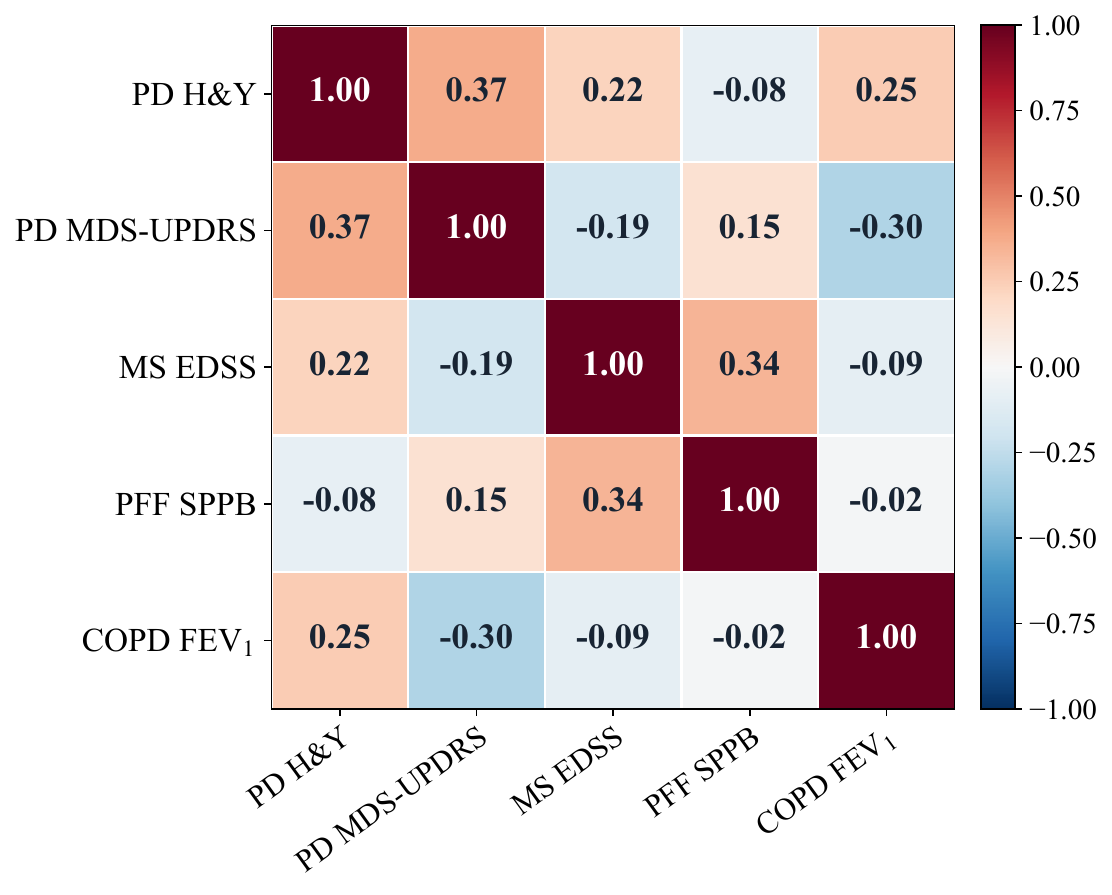}
	\caption{Mean complete relation matrix
		$\widetilde{\mathbf A}=\mathbf A+\mathbf I$ across the five matched splits.}
	\label{fig:learned_relation_matrix}
\end{wrapfigure}


\paragraph{Learned cross-objective relations.}
The optimiser learns the zero-diagonal cross-objective matrix $\mathbf A$. For
interpretation, we report the complete matrix
$\widetilde{\mathbf A}=\mathbf A+\mathbf I$, whose unit diagonal represents
each objective's relation with itself. Figure~\ref{fig:learned_relation_matrix}
shows its element-wise mean over the five matched splits. The strongest
positive relation is between the two PD outcomes ($0.37$), consistent with
their shared cohort and related measures of PD severity. Substantial positive
relations also emerge between MS EDSS and PFF SPPB ($0.34$), PD H\&Y and COPD
FEV$_1$ ($0.25$), and PD H\&Y and MS EDSS ($0.22$). These cross-disease
relations demonstrate that DeMMO can identify transferable mobility structure
even when cohorts share no participants, providing a mechanism for the gains
over the independently fitted cFSGL models.
The strongest negative relations are PD MDS--UPDRS--COPD FEV$_1$ ($-0.30$)
and PD MDS--UPDRS--MS EDSS ($-0.19$). Note that relation signs describe the orientation
of the learned longitudinal DMO effects; they do not represent associations
between diseases or correlations between raw clinical scores. This distinction
is especially important for COPD FEV$_1$, for which higher values indicate
better pulmonary function, whereas higher values of the other outcomes indicate
greater impairment. By learning both positive and negative relations, DeMMO
can exploit aligned as well as complementary cross-objective structure rather
than imposing uniformly positive information sharing.

\subsection{Longitudinal Stability Selection}
\label{sec:longitudinal_stability}

\begin{figure*}[!t]
	\centering
	\includegraphics[width=\textwidth]{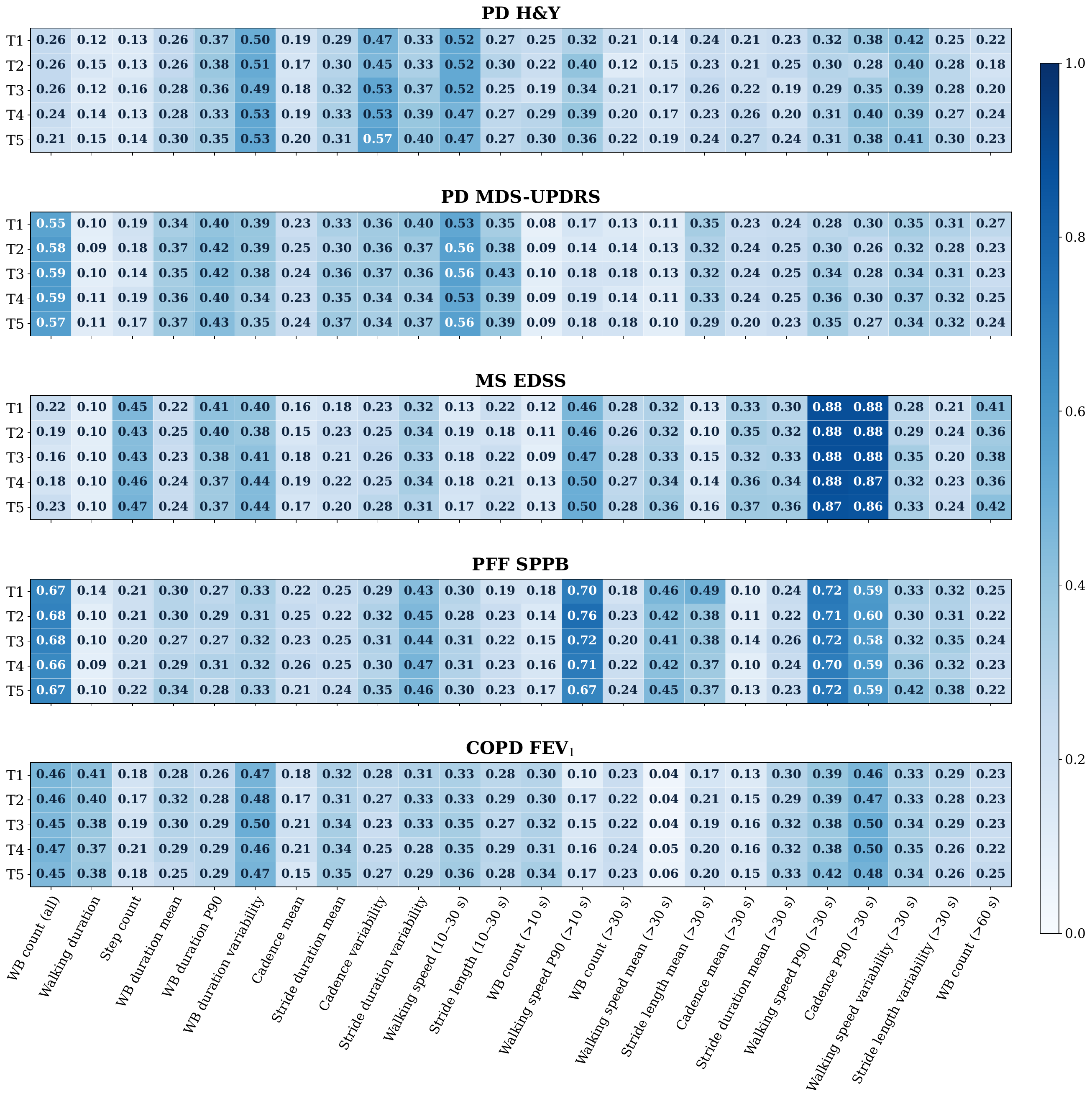}
	\caption{DMO selection probabilities across 100 half-sample fits (ten
		hyperparameter settings with ten repetitions) for each of the five
		prediction outcomes and visits T1--T5.}
	\label{fig:longitudinal_stability}
\end{figure*}

We perform longitudinal stability selection
\citep{zhou2024integrating,fan2026beyond,zhou2023efficient} over ten randomly
selected hyperparameter settings, each repeated on ten half-samples of the
training participants. Figure~\ref{fig:longitudinal_stability} reports how
often each DMO has a non-zero visit-specific coefficient in the resulting 100
fits. The profiles are broadly stable across visits but differ across
objectives. The most stable DMOs are walking-bout-duration and cadence
variability for PD H\&Y; total walking-bout count for PD MDS--UPDRS;
high-percentile walking speed and cadence for MS EDSS and PFF SPPB; and cadence
P90, walking-bout-duration variability, and walking-bout count for COPD
FEV$_1$. The strongest selection probabilities occur for MS EDSS, particularly
for walking-speed and cadence P90 in bouts longer than 30 seconds ($0.878$ and
$0.874$, respectively). PFF SPPB shows a related emphasis on high-percentile
pace and rhythm, whereas the two PD objectives favour partially overlapping
but distinct measures. Consistently, variation across objectives is
substantially greater than variation across visits ($0.111$ versus $0.019$),
and no DMO has a mean selection probability above $0.4$ for all five
objectives. This indicates that DMO relevance is longitudinally persistent but
not universal across diseases or clinical measurements.
These findings complement the learned relation matrix: stability selection identifies reliably selected DMOs, whereas the relation matrix captures the alignment of their longitudinal effects. Shared DMOs may therefore have either aligned or opposing effects across objectives. DeMMO thus identifies persistent, objective-specific mobility patterns while enabling selective information sharing. These patterns require further clinical validation.

A broader value of DeMMO is that its structured and interpretable longitudinal
DMO mappings enable reliable feature discovery and can subsequently support
concise explanations in monitoring software. As a proof of concept,
Appendix~\ref{sec:clinical_application} applies stability selection to the
DeMMO mappings and uses the two most stable MS DMOs to construct interpretable
binary and three-class EDSS rules. This illustrates how the proposed framework
can model all 24 DMOs while presenting users with a small and understandable
set of indicators associated with a detected mobility change. The application
remains exploratory and requires prospective validation.

\section{Conclusion}
\label{sec:con}
We presented DeMMO, an interpretable framework for jointly modelling
longitudinal DMOs across multiple diseases and clinical outcomes. By learning
signed relations between longitudinal DMO mappings, DeMMO supports selective
information sharing across participant-disjoint cohorts without requiring
paired observations. On four Mobilise-D disease cohorts, five prediction
objectives, and five visits, DeMMO achieved the best overall performance among
the evaluated methods and led in most outcome- and visit-specific comparisons.
The learned relations and stability analysis further show that informative DMO
patterns are persistent across visits yet distinct across clinical objectives,
supporting structured sharing rather than a universal mobility profile. Owing
to its simple and interpretable formulation, DeMMO provides both accurate
predictions and clinically inspectable longitudinal patterns. Future work will
validate these patterns in independent cohorts and examine their
generalisability to additional diseases, outcomes, and longer follow-up
periods.

\subsection*{AI use statement}
Generative AI tools were used to assist with language editing, LaTeX
formatting, and code review. The authors reviewed all AI-assisted material,
verified the reported claims and experimental results, and take responsibility
for the final content of the paper and its associated artifacts.

\subsection*{Reproducibility statement}
The paper specifies the model, optimisation procedure, data splitting,
preprocessing, hyperparameter selection, and evaluation protocol. Source code
and privacy-safe intermediate and aggregate results are publicly available at
\url{https://github.com/menghui-zhou/DeMMO}. Additional analyses are reported
in the appendix.

\bibliographystyle{iclr2027_conference}
\bibliography{menghui_total,menghui_total_DMO}

\newpage
\begin{center}
	\huge Appendix
\end{center}

\appendix

\section{Related Work}
\label{sec:related}
\subsection{Real-World DMO Estimation and Clinical Validation}
\label{sec:rw_dmo_validation}

Research on DMOs has examined whether mobility can be estimated accurately from wearable-sensor signals and whether the resulting measures are clinically meaningful \citep{goldsack2020verification,cerreta2020digital,polhemus2021walking}. Consensus definitions have improved comparability across sensors, algorithms, and walking-bout definitions \citep{kluge2021consensus}. The Mobilise-D Technical Validation Study further demonstrated the feasibility of deriving real-world DMOs from a single wearable device across several mobility-limiting conditions \citep{mazza2021technical,keogh2023acceptability}. Detection and cadence estimation were generally robust, whereas stride length and walking speed were more affected by context and impairment severity \citep{mico2023assessing,kirk2024mobilise}.
Clinical evidence remains dominated by known-groups comparisons and concurrent associations, with less support for predictive validity, responsiveness, and ecological validity \citep{polhemus2021walking}. Differences between laboratory and daily-life gait further motivate real-world validation \citep{shah2020laboratory,corra2021comparison}. Recent Mobilise-D studies have related individual DMOs to clinical measures and severity or recovery groups \citep{megaritis2026construct,eckert2026construct}, but do not show how multiple DMOs should be combined or how their relationships with clinical status evolve over time.

This literature reflects the distinction between technical and clinical
validation. Technical studies establish that a sensor and algorithm can
measure a walking construct with acceptable accuracy, whereas clinical
validation asks whether that construct is associated with an outcome of
interest and is fit for a specified context of use
\citep{goldsack2020verification,rochester2020roadmap}. Mobilise-D was designed
around this staged process: its technical validation study evaluated the
measurement pipeline across conditions, while the Clinical Validation Study
linked harmonised real-world DMOs to disease-specific clinical outcomes
\citep{mazza2021technical,mikolaizak2022connecting}. Construct-validity analyses
in COPD, PFF, and MS provide important outcome-specific evidence
\citep{megaritis2026construct,eckert2026construct,brittain2026testing}, but are
primarily cross-sectional. DeMMO complements these studies by analysing the
joint longitudinal contribution of the complete DMO panel rather than testing
each DMO separately.

\subsection{Longitudinal DMO Mining}
\label{sec:longitudinal_dmo_mining}

Longitudinal DMO research remains limited. Reviews in PD found few studies with extended follow-up and a literature dominated by active or hospital-based tests, patient--control comparisons, and cross-sectional associations, leaving longitudinal change and predictive utility underexplored \citep{rabano2025digital,kirk2025systematic}.
Emerging studies show that selected measures can capture change in PD, MS, and hip-fracture recovery \citep{mirelman2024digital,poleur2026stride,engdal2024real}. However, they are disease-specific and assess only a few prespecified outcomes; they do not jointly model how multivariate DMO--outcome relationships evolve across visits. To our knowledge, stable and visit-specific DMOs have not been systematically identified across multiple outcomes and mobility-limiting conditions. The Mobilise-D Clinical Validation Study provides the harmonised longitudinal data needed to address this gap \citep{mikolaizak2022connecting}.

Longitudinal evidence can address several distinct questions: whether a DMO
changes over time, whether its change tracks clinical progression, whether it
predicts a future endpoint, and whether its association with an endpoint is
stable across visits. These questions are not interchangeable. A measure may
show a population-level trend but contribute little to individual prediction,
or remain predictive even when its marginal mean changes little. Existing
studies have mainly examined trajectories of individual measures or their
association with a single clinical endpoint
\citep{mirelman2024digital,engdal2024real,poleur2026stride}. In contrast, our
setting estimates a visit-specific multivariate mapping from 24 correlated
DMOs to each clinical outcome. Stability selection then distinguishes effects
that persist across visits and participant subsamples from those that are
visit-specific, without interpreting statistical selection alone as clinical
validation.

Attrition and incomplete follow-up add a further difficulty. Later visits
often contain fewer observations, making independent visit-wise models
unstable, whereas complete-case trajectory analysis would discard
participants with partially observed follow-up. DeMMO retains all valid
participant--visit records and borrows strength through coefficient-level
temporal regularisation. It therefore targets the evolution of DMO--outcome
associations rather than requiring every participant to have a complete
five-visit trajectory.

\subsection{Multi-Task Learning for Longitudinal Modelling}
\label{sec:mtl_longitudinal_cross_disease}

Multi-task learning (MTL) \citep{zhang2021survey, ma2022kg, li2020multi} has been applied to disease progression by treating predictions at different future time points as related tasks. Studies in Alzheimer's disease use temporal smoothness, structured sparsity, and adaptive or robust regularisers to identify shared and time-specific biomarkers \citep{zhou2011multi,zhou2012modeling,zhou2022multi,zhou2023robust,zhou2024integrating}. These methods generally predict multiple future outcomes from baseline features \citep{zhou2011multi,zhou2012modeling,zhou2022model,zhou2024integrating,fan2026beyond}; in Mobilise-D, both DMOs and outcomes are repeatedly observed, so the goal is instead to model how their relationships evolve across visits.
Mobilise-D also spans multiple outcomes and diseases. Recent MTL studies often define outcome relationships through patient-level correlations within one disease \citep{fan2026beyond}; this is unsuitable for non-overlapping disease cohorts. Cross-disease relationships must instead be learned from shared DMO-based prediction structures while preserving disease-specific patterns.
Fused Lasso and sparse group Lasso provide mechanisms for temporal smoothness and structured selection \citep{tibshirani2005sparsity,zhou2012modeling,simon2013sparse}. Here, they identify stable and visit-specific DMOs among 24 measures for parsimony and clinical interpretability. Existing MTL methods do not jointly accommodate repeated DMO--outcome relationships, multiple outcomes, and non-overlapping disease cohorts, motivating a framework that learns temporal, outcome-specific, and cross-disease structures simultaneously.

The task definition is central to this distinction. In conventional
longitudinal MTL, tasks commonly correspond to future time points, and a
shared baseline representation is used to predict several follow-up outcomes
\citep{zhou2011multi,zhou2012modeling,zhou2022model}. Robust and adaptive
extensions allow task-specific deviations or learn relations among these time
points \citep{zhou2023robust,zhou2024integrating}. In DeMMO, each
disease--outcome pair is instead one prediction objective, and its coefficient
matrix contains the complete sequence of visit-specific DMO mappings. Temporal
dependence is modelled within each objective, while relation learning operates
between objectives. This separation prevents a strong temporal relation from
being confused with evidence that two clinical outcomes share the same DMO
structure.

Cross-outcome learning also differs from ordinary multi-output regression.
The two PD outcomes are observed in the same cohort, but MS, PFF, and COPD outcomes
come from different participants and have different clinical scales. Raw
outcome covariance is therefore unavailable across diseases and would not, in
any case, directly describe similarity between their DMO effects. DeMMO learns
a symmetric signed relation graph from the longitudinal coefficient mappings.
Positive and negative weights can represent aligned and oppositely oriented
DMO patterns, while the zero diagonal excludes trivial self-relations. This
parameter-level mechanism enables selective sharing across non-overlapping
cohorts without pooling participant records or assuming identical effects.

\subsection{Deep Learning for Disease Progression Modelling}
\label{sec:deep_disease_modelling}

Deep learning has been widely applied to large-scale electronic health records, medical images, and physiological signals. Representative models learn patient representations or temporal patterns to predict diseases, diagnoses, medications, and future conditions \citep{miotto2016deep,choi2016doctor,li2020behrt,bai2018interpretable}. Recurrent models have also jointly predicted longitudinal imaging, diagnoses, cognitive scores, and ventricular volume in Alzheimer's disease while accommodating missing observations \citep{ghazi2019training,nguyen2020predicting}.
These methods can capture nonlinear temporal dependencies but generally benefit from high-dimensional inputs, dense observations, or large patient populations. Mobilise-D instead contains 24 structured DMOs, at most five visits per participant, and smaller disease- and outcome-specific samples. Its main challenge is to identify interpretable, time-varying DMO relationships across clinical outcomes and diseases. A structured multi-task model therefore provides more appropriate inductive biases and more interpretable effects than a high-capacity deep sequence model.

Deep models for irregular clinical time series can explicitly encode missing
values, elapsed time, or attention across observations
\citep{che2018recurrent,shukla2021multi}. Such architectures are valuable when
the principal signal lies in dense, heterogeneous event sequences. Here,
however, the inputs are a small, clinically defined DMO panel observed on a
fixed five-visit schedule, and the scientific objective includes identifying
which DMO is relevant at which visit. Post-hoc feature attribution would not
provide the same direct coefficient trajectories or structured sparsity.

We nevertheless include neural regression and continuous-label representation
learning baselines in the empirical comparison. This tests whether nonlinear
capacity or label-aware embeddings improve prediction in the available sample
regime. The comparison is therefore not intended to claim that structured
linear models dominate deep learning generally; rather, it evaluates whether
the temporal, sparse, and cross-objective inductive biases of DeMMO are better
matched to this particular longitudinal clinical problem.

\section{Definitions of the 24 Digital Mobility Outcomes}
\label{sec:dmo_definitions}
Table~\ref{tab:dmo_definitions} defines the 24 weekly aggregated DMOs used as
predictors. A walking bout (WB) is a continuous period of detected walking.
In the variable names, \texttt{all}, \texttt{10}, \texttt{30}, and
\texttt{60} denote all WBs or WBs longer than 10, 30, and 60 seconds,
respectively; \texttt{1030} denotes WBs lasting 10--30 seconds. The suffixes
\texttt{avg}, \texttt{p90}, \texttt{var}, and \texttt{sum} denote the mean,
90th percentile, variability, and amount/count aggregation, while
\texttt{w} indicates aggregation over the monitoring week.

\begin{table*}[!t]
\centering
\caption{Definitions of the 24 weekly DMO predictors. WB denotes walking bout.}
\label{tab:dmo_definitions}
\scriptsize
\begin{tabular}{p{0.18\textwidth}p{0.26\textwidth}p{0.18\textwidth}p{0.26\textwidth}}
\toprule
Variable & Meaning & Variable & Meaning \\
\midrule
\texttt{wb\_all\_sum\_w} & Number of WBs &
\texttt{wb\_10\_sum\_w} & Number of WBs longer than 10 s \\
\texttt{walkdur\_all\_sum\_w} & Total walking duration &
\texttt{ws\_10\_p90\_w} & 90th percentile of walking speed in WBs longer than 10 s \\
\texttt{wbsteps\_all\_sum\_w} & Total number of walking steps &
\texttt{wb\_30\_sum\_w} & Number of WBs longer than 30 s \\
\texttt{wbdur\_all\_avg\_w} & Mean WB duration &
\texttt{ws\_30\_avg\_w} & Mean walking speed in WBs longer than 30 s \\
\texttt{wbdur\_all\_p90\_w} & 90th percentile of WB duration &
\texttt{strlen\_30\_avg\_w} & Mean stride length in WBs longer than 30 s \\
\texttt{wbdur\_all\_var\_w} & Variability of WB duration &
\texttt{cadence\_30\_avg\_w} & Mean cadence in WBs longer than 30 s \\
\texttt{cadence\_all\_avg\_w} & Mean cadence across all WBs &
\texttt{strdur\_30\_avg\_w} & Mean stride duration in WBs longer than 30 s \\
\texttt{strdur\_all\_avg\_w} & Mean stride duration across all WBs &
\texttt{ws\_30\_p90\_w} & 90th percentile of walking speed in WBs longer than 30 s \\
\texttt{cadence\_all\_var\_w} & Variability of cadence across all WBs &
\texttt{cadence\_30\_p90\_w} & 90th percentile of cadence in WBs longer than 30 s \\
\texttt{strdur\_all\_var\_w} & Variability of stride duration across all WBs &
\texttt{ws\_30\_var\_w} & Variability of walking speed in WBs longer than 30 s \\
\texttt{ws\_1030\_avg\_w} & Mean walking speed in WBs lasting 10--30 s &
\texttt{strlen\_30\_var\_w} & Variability of stride length in WBs longer than 30 s \\
\texttt{strlen\_1030\_avg\_w} & Mean stride length in WBs lasting 10--30 s &
\texttt{wb\_60\_sum\_w} & Number of WBs longer than 60 s \\
\bottomrule
\end{tabular}
\end{table*}

Together, these variables characterise five complementary aspects of
real-world mobility: walking amount, bout pattern, pace, rhythm, and
variability. Bout-duration thresholds distinguish short habitual walking from
longer, sustained walking, while mean, upper-percentile, and variability
summaries capture typical, high-performance, and within-person heterogeneous
mobility behaviour, respectively.

\section{Optimization}
\label{sec:optimization}
The unified objective in Equation \ref{eq:dummo_objective} is not jointly convex in the longitudinal coefficient matrices and relation matrix because of the multiplicative term \(\mathbf{U}\mathbf{A}\). However, it is convex in either block when the other is fixed. We therefore use alternating optimisation to update the longitudinal coefficient matrices \(\{\mathbf{W}_{d,q}\}\) with \(\mathbf{A}\) fixed, followed by the symmetric relation matrix \(\mathbf{A}\) with \(\{\mathbf{W}_{d,q}\}\) fixed.

For simplicity, we assign each disease--outcome pair a unique index \(m\in\{1,\ldots,M\}\), where \(M=5\), and denote its coefficient matrix by \(\mathbf{W}_m=[\mathbf{w}_{m1},\ldots,\mathbf{w}_{mT}]\in\mathbb{R}^{p\times T}\). The corresponding data at visit \(t\) are denoted by \(\mathbf{X}_{mt}\), \(\mathbf{y}_{mt}\), and \(n_{mt}\).

\subsection{Updating the Longitudinal Coefficient Matrices}
\label{sec:update_w}

For fixed \(\mathbf{A}\), the coefficient subproblem is decomposed into the smooth component
\begin{equation}
	\begin{aligned}
		f(\{\mathbf{W}_m\};\mathbf{A})
		=
		&
		\sum_{m=1}^{M}
		\sum_{t=1}^{T}
		\frac{1}{2n_{mt}}
		\left\|
		\mathbf{y}_{mt}
		-
		\mathbf{X}_{mt}\mathbf{w}_{mt}
		\right\|_2^2
		\\
		&+
		\frac{\lambda_{\mathrm{R}}}{2}
		\left\|
		\mathbf{U}-\mathbf{U}\mathbf{A}
		\right\|_F^2,
	\end{aligned}
\end{equation}
and the non-smooth component
\begin{equation}
	g(\{\mathbf{W}_m\})=\sum_{m=1}^{M}\left[\lambda_{\mathrm{FL}}\|\mathbf{W}_m\mathbf{R}^{\mathsf T}\|_1+\lambda_{\mathrm{L}}\|\mathbf{W}_m\|_1+\lambda_{\mathrm{G}}\|\mathbf{W}_m\|_{2,1}\right].
\end{equation}
The relation penalty \(\lambda_{\mathrm{A}}\|\mathbf{A}\|_F^2/2\) is constant in this subproblem and is therefore omitted. We minimise \(f+g\) using accelerated proximal gradient with backtracking line search.

The regression gradient for objective \(m\) at visit \(t\) is
\begin{equation}
	\nabla_{\mathbf{w}_{mt}}f_{\mathrm{reg}}=\frac{1}{n_{mt}}\mathbf{X}_{mt}^{\mathsf T}\left(\mathbf{X}_{mt}\mathbf{w}_{mt}-\mathbf{y}_{mt}\right).
\end{equation}

For the relation term, define \(\mathbf{B}=\mathbf{I}-\mathbf{A}\). Then \(f_{\mathrm{rel}}=\lambda_{\mathrm{R}}\|\mathbf{U}\mathbf{B}\|_F^2/2\), with gradient
\begin{equation}
	\nabla_{\mathbf{U}}f_{\mathrm{rel}}=\lambda_{\mathrm{R}}\mathbf{U}\mathbf{B}\mathbf{B}^{\mathsf T}.
\end{equation}
Because column \(m\) of \(\mathbf{U}\) is the vectorised form of \(\mathbf{W}_m\), its relation gradient is reshaped into a \(p\times T\) matrix. The complete gradient is therefore
\begin{equation}
	\nabla_{\mathbf{W}_m}f=\nabla_{\mathbf{W}_m}f_{\mathrm{reg}}+\operatorname{unvec}_{p\times T}\left([\nabla_{\mathbf{U}}f_{\mathrm{rel}}]_{:m}\right).
\end{equation}

Let \(\mathbf{Z}_m^{(k)}\) be the extrapolated coefficient matrix at inner iteration \(k\), and let \(L_k\) be the current estimate of the Lipschitz constant \citep{nesterov1983method, nesterov2003introductory}. The proximal-gradient step is
\begin{equation}
	\mathbf{V}_m^{(k)}=\mathbf{Z}_m^{(k)}-\frac{1}{L_k}\nabla_{\mathbf{W}_m}f(\{\mathbf{Z}_m^{(k)}\};\mathbf{A}),
\end{equation}
followed by
\begin{equation}
	\mathbf{W}_m^{(k+1)}=\operatorname{prox}_{g_m/L_k}\left(\mathbf{V}_m^{(k)}\right).
\end{equation}

\paragraph{Composite proximal operator.}

The coefficient update involves a composite penalty consisting of fused Lasso, element-wise Lasso, and group Lasso terms. Although proximal operators cannot generally be composed in an arbitrary order, this fused sparse group-Lasso penalty admits an exact and efficient decomposition \citep{zhou2023robust, zhou2012modeling}.

Specifically, the proximal problem is separable across prediction objectives and DMO rows. For objective \(m\) and DMO \(j\), let \(\mathbf{v}_{mj}\in\mathbb{R}^{T}\) denote the \(j\)-th row of \(\mathbf{V}_m^{(k)}\). The corresponding update is
\begin{equation}
	\begin{aligned}
		\mathbf{w}_{mj}^{(k+1)}
		=
		\arg\min_{\mathbf{w}\in\mathbb{R}^{T}}
		\frac{1}{2}\|\mathbf{w}-\mathbf{v}_{mj}\|_2^2
		+
		\frac{\lambda_{\mathrm{L}}}{L_k}\|\mathbf{w}\|_1+
		\frac{\lambda_{\mathrm{FL}}}{L_k}\|\mathbf{R}\mathbf{w}\|_1
		+
		\frac{\lambda_{\mathrm{G}}}{L_k}\|\mathbf{w}\|_2.
	\end{aligned}
\end{equation}

This problem is solved exactly in two stages. First, the fused and element-wise sparse solution is obtained from
\begin{equation}
	\begin{aligned}
		\widetilde{\mathbf{w}}_{mj}
		=
		\arg\min_{\mathbf{w}\in\mathbb{R}^{T}}
		\frac{1}{2}
		\|\mathbf{w}-\mathbf{v}_{mj}\|_2^2
		+
		\frac{\lambda_{\mathrm{L}}}{L_k}
		\|\mathbf{w}\|_1
+
		\frac{\lambda_{\mathrm{FL}}}{L_k}
		\|\mathbf{R}\mathbf{w}\|_1.
	\end{aligned}
\end{equation}
Second, row-wise group shrinkage is applied:
\begin{equation}
	\mathbf{w}_{mj}^{(k+1)}=\left(1-\frac{\lambda_{\mathrm{G}}}{L_k\|\widetilde{\mathbf{w}}_{mj}\|_2}\right)_{+}\widetilde{\mathbf{w}}_{mj},
\end{equation}
where \((a)_{+}=\max(a,0)\). The first stage identifies temporally fused and visit-specific coefficients, while the second removes the complete DMO trajectory when its magnitude is insufficient. This decomposition exactly evaluates the composite proximal operator and can be implemented using established fast fused-Lasso algorithms.

Backtracking line search is used to select \(L_k\), and Nesterov acceleration is applied after each accepted update. The inner iterations terminate when the relative change in the coefficient-block objective falls below a prescribed tolerance.

\subsection{Updating the Symmetric Relation Matrix}
\label{sec:update_a}

For fixed coefficient matrices, \(\mathbf{U}\) is constant and the relation subproblem becomes
\begin{equation}
	\begin{aligned}
		\min_{\mathbf{A}}
		\quad
		&
		\frac{\lambda_{\mathrm{R}}}{2}
		\left\|
		\mathbf{U}-\mathbf{U}\mathbf{A}
		\right\|_F^2
		+
		\frac{\lambda_{\mathrm{A}}}{2}
		\left\|
		\mathbf{A}
		\right\|_F^2
		\\
		\text{\rm s.t.}\quad
		&
		\mathbf{A}=\mathbf{A}^{\mathsf T},
		\qquad
		\operatorname{diag}(\mathbf{A})=\mathbf{0}.
	\end{aligned}
\end{equation}
This is a strongly convex quadratic problem because \(\lambda_{\mathrm{A}}>0\). To satisfy symmetry and the zero-diagonal constraint exactly, we parameterise only the \(K=M(M-1)/2\) upper-triangular entries.

For each pair \(i<j\), define \(\mathbf{E}_{ij}=\mathbf{e}_i\mathbf{e}_j^{\mathsf T}+\mathbf{e}_j\mathbf{e}_i^{\mathsf T}\), where \(\mathbf{e}_i\) is the \(i\)-th standard basis vector in \(\mathbb{R}^{M}\). Any feasible relation matrix can then be written as \(\mathbf{A}(\mathbf{a})=\sum_{i<j}a_{ij}\mathbf{E}_{ij}\), where \(\mathbf{a}\in\mathbb{R}^{K}\).
Let \(\mathbf{b}=\operatorname{vec}(\mathbf{U})\) and define
\begin{equation}
	\mathbf{C}
	=
	\left[
	\operatorname{vec}(\mathbf{U}\mathbf{E}_{12}),
	\operatorname{vec}(\mathbf{U}\mathbf{E}_{13}),
	\ldots,
	\operatorname{vec}(\mathbf{U}\mathbf{E}_{M-1,M})
	\right].
\end{equation}

Since \(\operatorname{vec}(\mathbf{U}\mathbf{A}(\mathbf{a}))=\mathbf{C}\mathbf{a}\) and \(\|\mathbf{A}(\mathbf{a})\|_F^2=2\|\mathbf{a}\|_2^2\), the constrained matrix problem reduces to
\begin{equation}
	\min_{\mathbf{a}\in\mathbb{R}^{K}}
	\quad
	\frac{\lambda_{\mathrm{R}}}{2}
	\left\|
	\mathbf{b}-\mathbf{C}\mathbf{a}
	\right\|_2^2
	+
	\lambda_{\mathrm{A}}
	\left\|
	\mathbf{a}
	\right\|_2^2.
\end{equation}
Its unique minimiser is obtained from
\begin{equation}
	\left(\lambda_{\mathrm{R}}\mathbf{C}^{\mathsf T}\mathbf{C}+2\lambda_{\mathrm{A}}\mathbf{I}\right)\mathbf{a}=\lambda_{\mathrm{R}}\mathbf{C}^{\mathsf T}\mathbf{b}.
\end{equation}

The relation matrix is then formed as \(\mathbf{A}=\sum_{i<j}a_{ij}\mathbf{E}_{ij}\). This parameterisation satisfies both constraints by construction. In this study, \(M=5\) and hence \(K=10\), so the update can be computed efficiently using a standard linear-system solver.

\subsection{Alternating Optimisation Algorithm}
\label{sec:alternating_algorithm}

The complete optimisation procedure is summarised in
Algorithm~\ref{alg:alternating_optimization}. We initialise the relation
matrix as $\mathbf{A}^{(0)}=\mathbf{0}$. The coefficient matrices may be
initialised either at zero or using independently fitted visit-specific
regression models. During subsequent outer iterations, each coefficient
subproblem is warm-started from the solution obtained in the preceding
iteration.

\begin{algorithm}[!t]
	\caption{Alternating Optimisation of the Longitudinal Coefficient Matrices
		and Cross-Objective Relation Matrix.}
	\label{alg:alternating_optimization}
	\begin{algorithmic}[1]
		\REQUIRE Training data
		$\{\mathbf{X}_{mt},\mathbf{y}_{mt}\}_{m=1,t=1}^{M,T}$;
		regularisation parameters
		$\lambda_{\mathrm{FL}},\lambda_{\mathrm{L}},
		\lambda_{\mathrm{G}},\lambda_{\mathrm{R}},
		\lambda_{\mathrm{A}}$;
		maximum number of outer iterations $R_{\max}$
		
		\STATE Initialise $\{\mathbf{W}_m^{(0)}\}_{m=1}^{M}$ and
		$\mathbf{A}^{(0)}=\mathbf{0}$
		
		\FOR{$r=0,\ldots,R_{\max}-1$}
		
		\STATE Fix $\mathbf{A}^{(r)}$ and update
		$\{\mathbf{W}_m^{(r+1)}\}_{m=1}^{M}$ using accelerated
		proximal-gradient iterations with backtracking line search
		
		\STATE Evaluate the composite proximal operator row-wise using
		fused-Lasso signal approximation followed by group shrinkage
		
		\STATE Construct $\mathbf{U}^{(r+1)}$ by vectorising
		$\{\mathbf{W}_m^{(r+1)}\}_{m=1}^{M}$
		
		\STATE Construct $\mathbf{C}^{(r+1)}$ and
		$\mathbf{b}^{(r+1)}$ from $\mathbf{U}^{(r+1)}$
		
		\STATE Solve
		\[
		\left(
		\lambda_{\mathrm{R}}
		\mathbf{C}^{\mathsf T}\mathbf{C}
		+
		2\lambda_{\mathrm{A}}\mathbf{I}
		\right)\mathbf{a}
		=
		\lambda_{\mathrm{R}}
		\mathbf{C}^{\mathsf T}\mathbf{b}
		\]
		
		\STATE Form the symmetric, zero-diagonal relation matrix
		$\mathbf{A}^{(r+1)}$ from $\mathbf{a}$
		
		\IF{the outer convergence criterion is satisfied}
		\STATE \textbf{break}
		\ENDIF
		
		\ENDFOR
		
		\ENSURE $\{\mathbf{W}_m\}_{m=1}^{M}$ and $\mathbf{A}$
	\end{algorithmic}
\end{algorithm}

The outer iterations terminate when the relative change in the complete
objective satisfies
\begin{equation}
	\frac{
		\left|
		\mathcal{J}^{(r+1)}-\mathcal{J}^{(r)}
		\right|
	}{
		\max\left(1,\left|\mathcal{J}^{(r)}\right|\right)
	}
	\leq \varepsilon,
\end{equation}
where $\mathcal{J}^{(r)}$ denotes the unified objective value at outer
iteration $r$. A maximum number of outer iterations is imposed as an
additional safeguard.

\subsubsection{Convergence and Computational Complexity}
\label{sec:optimization_complexity}

For fixed $\mathbf{A}$, the coefficient subproblem is convex but non-smooth.
Because its composite proximal operator can be computed exactly, the
accelerated proximal-gradient procedure with backtracking converges to the
global minimiser of this subproblem. For fixed coefficient matrices, the
relation-matrix subproblem is strongly convex when
$\lambda_{\mathrm{A}}>0$ and therefore has a unique minimiser.

If both block subproblems are solved exactly, or to sufficient numerical
accuracy, each outer iteration does not increase the unified objective.
Because the objective is bounded below, the sequence of objective values
converges. Under standard regularity conditions, every accumulation point of
the alternating sequence is a block-coordinate stationary point. However,
the complete objective is biconvex rather than jointly convex because of the
interaction between $\mathbf{U}$ and $\mathbf{A}$. Convergence to a global
minimiser is therefore not guaranteed and may depend on the initialisation.

For the coefficient update, evaluating the regression gradients across all
objectives and visits requires
\begin{equation*}
	\mathcal{O}\left(
	p\sum_{m=1}^{M}\sum_{t=1}^{T}n_{mt}
	\right)
\end{equation*}
operations. The relation-gradient term involves multiplication between the
$pT\times M$ coefficient representation and the $M\times M$ relation
matrices, requiring $\mathcal{O}\left(pTM^{2}\right)$
operations.

The composite proximal operator is separable across the $Mp$ longitudinal
coefficient trajectories. When a linear-time one-dimensional fused-Lasso
solver is used, its overall cost is
\begin{equation*}
	\mathcal{O}\left(MpT\right).
\end{equation*}
Thus, if $I_{\mathrm{in}}$ accelerated proximal-gradient iterations are
required, the coefficient-update cost per outer iteration is approximately
\begin{equation}
	\mathcal{O}\left[
	I_{\mathrm{in}}
	\left(
	p\sum_{m=1}^{M}\sum_{t=1}^{T}n_{mt}
	+
	pTM^{2}
	+
	MpT
	\right)
	\right].
\end{equation}

The relation update contains $K=M(M-1)/2$ free edge variables. Constructing
its quadratic system depends on the $pT$-dimensional representations of the
$M$ objectives, while solving the resulting $K\times K$ linear system by a
direct method requires $\mathcal{O}(K^{3})$ operations. Because $K$ depends
only on the number of prediction objectives, this update remains inexpensive
when $M$ is small.

In the present study, $p=24$, $T=5$, $M=5$, and $K=10$. Consequently, the
composite proximal operations and relation-matrix update are small, and the
overall computational cost is dominated by repeated regression-gradient
evaluations across participants, visits, and prediction objectives.

\section{Non-Convex Extensions of DeMMO}
\label{sec:noncon}

The fused-Lasso, Lasso, and group-Lasso regularisers in DeMMO provide a
convex and computationally tractable formulation for temporal smoothing and
DMO selection. However, convex sparsity penalties may over-shrink the
coefficients of informative DMOs, particularly when their effects are weak,
correlated, or confined to specific visits, thereby introducing estimation
bias \citep{fan2001variable}. Such shrinkage may reduce both predictive
performance and the ability to recover clinically meaningful
longitudinal patterns.

Motivated by non-convex extensions of the fused sparse-group Lasso
\citep{zhou2012modeling}, we introduce two variants, DeMMO-var1 and DeMMO-var2.
Both use a concave square-root penalty to reduce shrinkage of sufficiently
strong DMO effects while retaining sparsity. They differ in how DMO selection
and temporal smoothness are organised. DeMMO-var1 treats them as separate
components, allowing their strengths to be controlled independently.
DeMMO-var2 couples them within a single DMO-level penalty, such that retention
of a DMO depends jointly on its coefficient magnitude and temporal
variation. Comparing these variants allows us to examine whether longitudinal
DMO modelling benefits more from independent or coupled regularisation.

For concise notation, let
$\mathbf{w}_{d,q,j:}\in\mathbb{R}^{T}$ denote the coefficients of DMO $j$
across all visits for disease $d$ and outcome $q$. Both variants retain the
longitudinal regression loss $\mathcal{L}(\mathcal{W})$ and the
relation-learning regulariser
\begin{equation}
	\mathcal{R}_{\mathrm{rel}}(\mathbf{U},\mathbf{A})
	=
	\frac{\lambda_{\mathrm{R}}}{2}
	\|\mathbf{U}-\mathbf{U}\mathbf{A}\|_F^2
	+
	\frac{\lambda_{\mathrm{A}}}{2}
	\|\mathbf{A}\|_F^2.
\end{equation}

\subsection{DeMMO-var1: Separate Selection and Temporal Smoothing}

DeMMO-var1 replaces the Lasso and group-Lasso penalties with a composite
$\ell_{(0.5,1)}$ penalty while retaining fused temporal regularisation as a
separate component:
\begin{equation}
	\begin{aligned}
		\min_{\mathcal{W},\mathbf{A}}\quad
		&\mathcal{L}(\mathcal{W})
		+\lambda_{\mathrm{NC}}\sum_{d,q,j}
		\sqrt{\|\mathbf{w}_{d,q,j:}\|_1}
		+\lambda_{\mathrm{FL}}\sum_{d,q}
		\|\mathbf{W}_{d,q}\mathbf{R}^{\mathsf T}\|_1
		+\mathcal{R}_{\mathrm{rel}}(\mathbf{U},\mathbf{A}) \\
		\text{\rm s.t.}\quad
		&\mathbf{A}=\mathbf{A}^{\mathsf T},
		\qquad
		\operatorname{diag}(\mathbf{A})=\mathbf{0}.
	\end{aligned}
	\label{eq:dummo_var1}
\end{equation}
The outer square root promotes DMO-level sparsity, whereas the inner
$\ell_1$ norm permits visit-specific coefficients to be removed. Temporal
smoothness is controlled independently by $\lambda_{\mathrm{FL}}$.

\subsection{DeMMO-var2: Coupled Selection and Temporal Smoothing}

DeMMO-var2 instead incorporates coefficient sparsity and temporal variation
within the same DMO-level penalty. Define
\begin{equation}
	\phi_{d,q,j}(\mathcal{W})
	=
	\|\mathbf{w}_{d,q,j:}\mathbf{R}^{\mathsf T}\|_1
	+\beta\|\mathbf{w}_{d,q,j:}\|_1,
\end{equation}
where $\beta$ controls their relative contributions. The objective becomes
\begin{equation}
	\begin{aligned}
		\min_{\mathcal{W},\mathbf{A}}\quad
		&\mathcal{L}(\mathcal{W})
		+\lambda_{\mathrm{NC}}
		\sum_{d,q,j}\sqrt{\phi_{d,q,j}(\mathcal{W})}
		+\mathcal{R}_{\mathrm{rel}}(\mathbf{U},\mathbf{A})\\
		\text{\rm s.t.}\quad
		&\mathbf{A}=\mathbf{A}^{\mathsf T},
		\qquad
		\operatorname{diag}(\mathbf{A})=\mathbf{0}.
	\end{aligned}
	\label{eq:dummo_var2}
\end{equation}
Unlike DeMMO-var1, DeMMO-var2 determines whether to retain a DMO according to
both the magnitude and temporal variation of its longitudinal coefficients.

\subsection{Optimisation}

Unlike convex DeMMO, the two variants require an additional difference-of-
convex outer loop. We use the iterative reweighting procedure established for
non-convex fused sparse-group Lasso models
\citep{zhou2012modeling,zhou2023robust}. For a non-negative quantity $s$, the
concavity of the square root gives the majoriser
\begin{equation}
    \sqrt{s+\epsilon}
    \leq
    \sqrt{s^{(k)}+\epsilon}
    +\frac{s-s^{(k)}}{2\sqrt{s^{(k)}+\epsilon}},
    \label{eq:sqrt_majoriser}
\end{equation}
where equality holds at $s=s^{(k)}$. Thus, at reweighting iteration $k$, the
non-convex penalties are replaced by weighted convex $\ell_1$ and fused-Lasso
penalties. Define
\begin{equation}
	a_{d,q,j}^{(k)}
	=
	\frac{\lambda_{\mathrm{NC}}}
	{2\sqrt{\|\mathbf{w}_{d,q,j:}^{(k)}\|_1+\epsilon}},
	\qquad
	b_{d,q,j}^{(k)}
	=
	\frac{\lambda_{\mathrm{NC}}}
	{2\sqrt{\phi_{d,q,j}(\mathcal{W}^{(k)})+\epsilon}},
	\label{eq:nonconvex_weights}
\end{equation}
where $\epsilon>0$ prevents unbounded weights. Ignoring terms independent of
$\mathcal{W}$, the coefficient majorisers for DeMMO-var1 and DeMMO-var2 are,
respectively,
\begin{align}
    \mathcal{Q}_{\mathrm{var1}}^{(k)}
    =&\;\mathcal{L}(\mathcal{W})
    +\sum_{d,q,j}a_{d,q,j}^{(k)}
    \|\mathbf{w}_{d,q,j:}\|_1
    +\lambda_{\mathrm{FL}}\sum_{d,q,j}
    \|\mathbf{w}_{d,q,j:}\mathbf{R}^{\mathsf T}\|_1
    +\mathcal{R}_{\mathrm{rel}},
    \label{eq:var1_majoriser}\\
    \mathcal{Q}_{\mathrm{var2}}^{(k)}
    =&\;\mathcal{L}(\mathcal{W})
    +\sum_{d,q,j}b_{d,q,j}^{(k)}
    \left(
    \|\mathbf{w}_{d,q,j:}\mathbf{R}^{\mathsf T}\|_1
    +\beta\|\mathbf{w}_{d,q,j:}\|_1
    \right)
    +\mathcal{R}_{\mathrm{rel}}.
    \label{eq:var2_majoriser}
\end{align}
Both are weighted convex fused-Lasso problems. For fixed $\mathbf{A}$, we
solve the coefficient block using accelerated proximal-gradient iterations
with backtracking. The proximal step is separable across DMO trajectories and
uses the same one-dimensional fused-Lasso solver as convex DeMMO, but with the
row-specific weights in Eq.~(\ref{eq:nonconvex_weights}). For fixed
$\mathcal{W}$, the symmetric relation matrix is updated using the same
closed-form quadratic subproblem described in
Section~\ref{sec:update_a}.

Algorithm~\ref{alg:nonconvex_dummo} summarises the complete procedure. We
initialise both variants from the fitted convex DeMMO solution. This warm start
provides a stable coefficient pattern and avoids starting the reweighting
scheme near the singular point of the square-root penalty.

\begin{algorithm}[!t]
\caption{Non-Convex DeMMO Optimisation.}
\label{alg:nonconvex_dummo}
\begin{algorithmic}[1]
\REQUIRE Longitudinal data, variant parameters, $\epsilon$, and tolerances
\STATE Initialise $(\mathcal{W}^{(0)},\mathbf{A}^{(0)})$ from convex DeMMO
\FOR{$k=0,1,\ldots,K_{\mathrm{DC}}-1$}
    \STATE Compute $a_{d,q,j}^{(k)}$ for var1 or
    $b_{d,q,j}^{(k)}$ for var2 using Eq.~(\ref{eq:nonconvex_weights})
    \REPEAT
        \STATE Update $\mathcal{W}$ by accelerated proximal gradient on
        Eq.~(\ref{eq:var1_majoriser}) or Eq.~(\ref{eq:var2_majoriser})
        \STATE Update the symmetric zero-diagonal $\mathbf{A}$ by solving
        the relation-learning linear system
    \UNTIL{the convex majoriser satisfies the block-convergence criterion}
    \STATE Evaluate the original non-convex objective
    \IF{its relative change is below $\varepsilon_{\mathrm{DC}}$}
        \STATE \textbf{break}
    \ENDIF
\ENDFOR
\ENSURE $\mathcal{W}$ and $\mathbf{A}$
\end{algorithmic}
\end{algorithm}

The majorisation step is tight and each inner block update decreases its
convex surrogate. Consequently, the original objective is non-increasing and
the procedure converges to a stationary point under the usual boundedness and
accurate-subproblem assumptions. Global optimality is not guaranteed because
the objectives remain non-convex.

\paragraph{Computational cost.}
Let $K_{\mathrm{DC}}$ be the number of reweighting iterations,
$K_{\mathrm{B}}$ the number of alternating block updates per majoriser, and
$K_{\mathrm{PG}}$ the number of proximal-gradient iterations.
One gradient/proximal update has the same leading cost as convex DeMMO,
\begin{equation}
    \mathcal{O}\!\left(
    p\sum_{m=1}^{M}\sum_{t=1}^{T}n_{mt}
    +pTM^{2}+MpT
    \right).
\end{equation}
The complete coefficient cost is therefore
\begin{equation}
    \mathcal{O}\!\left[
    K_{\mathrm{DC}}K_{\mathrm{B}}K_{\mathrm{PG}}
    \left(
    p\sum_{m,t}n_{mt}+pTM^{2}+MpT
    \right)
    \right],
\end{equation}
in addition to the small relation-matrix solves. Convex DeMMO requires only
one alternating optimisation sequence, whereas each non-convex variant solves
a sequence of weighted convex DeMMO-like problems. In our implementation,
$K_{\mathrm{DC}}$, $K_{\mathrm{B}}$, and $K_{\mathrm{PG}}$ are capped at 15,
12, and 300, respectively, with early stopping at every level. The convex warm
start substantially reduces the realised number of iterations, but the two
variants remain more computationally expensive. Their additional predictive
flexibility therefore comes at the cost of longer model selection and
training, which is the principal trade-off examined in our experiments.

\section{Experimental Evaluation of the Non-Convex Variants}
\label{sec:nonconvex_variant_results}

Table~\ref{tab:nonconvex_variant_results} reports the results of DeMMO-var1 and
DeMMO-var2. The corresponding results for the original DeMMO are reported in
Table~\ref{tab:mobilise_results_summary} and
Tables~\ref{tab:visit_rmse_pd_hy}--\ref{tab:visit_rmse_copd}.

\begin{table*}[!t]
	\centering
	\caption{Comparison of the two non-convex DeMMO variants over five matched
		participant splits. Values are mean $\pm$ standard deviation. T1--T5 report
		visit-specific RMSE. The best mean between the two variants is shown
		in bold.}
	\label{tab:nonconvex_variant_results}
	\resizebox{\textwidth}{!}{%
		\begin{tabular}{llccccccc}
			\toprule
			Objective & Method & nMSE $\downarrow$ & wR $\uparrow$ & T1 & T2 & T3 & T4 & T5 \\
			\midrule
			\multirow{2}{*}{Overall}
			& DeMMO-var1 & 0.773 \spm0.035 & 0.459 \spm0.040 & -- & -- & -- & -- & -- \\
			& DeMMO-var2 & \cellcolor{blue!10}{\textbf{0.772 \spm0.033}} & \cellcolor{blue!10}{\textbf{0.459 \spm0.038}} & -- & -- & -- & -- & -- \\
			\midrule
			\multirow{2}{*}{PD H\&Y}
			& DeMMO-var1 & 0.946 \spm0.039 & 0.291 \spm0.058 & 0.536 \spm0.036 & \cellcolor{blue!10}{\textbf{0.504 \spm0.058}} & 0.549 \spm0.043 & \cellcolor{blue!10}{\textbf{0.522 \spm0.078}} & 0.555 \spm0.072 \\
			& DeMMO-var2 & \cellcolor{blue!10}{\textbf{0.944 \spm0.036}} & \cellcolor{blue!10}{\textbf{0.291 \spm0.056}} & \cellcolor{blue!10}{\textbf{0.536 \spm0.035}} & 0.506 \spm0.058 & \cellcolor{blue!10}{\textbf{0.546 \spm0.039}} & 0.522 \spm0.077 & \cellcolor{blue!10}{\textbf{0.552 \spm0.069}} \\
			\midrule
			\multirow{2}{*}{PD MDS--UPDRS}
			& DeMMO-var1 & \cellcolor{blue!10}{\textbf{0.889 \spm0.051}} & \cellcolor{blue!10}{\textbf{0.364 \spm0.059}} & \cellcolor{blue!10}{\textbf{11.930 \spm0.475}} & 11.689 \spm0.106 & 12.107 \spm0.697 & 11.492 \spm0.787 & 11.519 \spm0.769 \\
			& DeMMO-var2 & 0.889 \spm0.048 & 0.358 \spm0.053 & 11.991 \spm0.533 & \cellcolor{blue!10}{\textbf{11.689 \spm0.159}} & \cellcolor{blue!10}{\textbf{12.064 \spm0.639}} & \cellcolor{blue!10}{\textbf{11.480 \spm0.764}} & \cellcolor{blue!10}{\textbf{11.497 \spm0.737}} \\
			\midrule
			\multirow{2}{*}{MS EDSS}
			& DeMMO-var1 & \cellcolor{blue!10}{\textbf{0.485 \spm0.054}} & \cellcolor{blue!10}{\textbf{0.727 \spm0.039}} & \cellcolor{blue!10}{\textbf{0.820 \spm0.082}} & 0.830 \spm0.058 & \cellcolor{blue!10}{\textbf{0.991 \spm0.061}} & 1.013 \spm0.033 & \cellcolor{blue!10}{\textbf{1.093 \spm0.086}} \\
			& DeMMO-var2 & 0.487 \spm0.049 & 0.727 \spm0.035 & 0.822 \spm0.078 & \cellcolor{blue!10}{\textbf{0.828 \spm0.058}} & 0.997 \spm0.063 & \cellcolor{blue!10}{\textbf{1.010 \spm0.046}} & 1.098 \spm0.090 \\
			\midrule
			\multirow{2}{*}{PFF SPPB}
			& DeMMO-var1 & \cellcolor{blue!10}{\textbf{0.547 \spm0.060}} & \cellcolor{blue!10}{\textbf{0.681 \spm0.044}} & \cellcolor{blue!10}{\textbf{1.997 \spm0.058}} & 2.213 \spm0.189 & 2.262 \spm0.221 & \cellcolor{blue!10}{\textbf{2.213 \spm0.327}} & \cellcolor{blue!10}{\textbf{2.221 \spm0.165}} \\
			& DeMMO-var2 & 0.550 \spm0.065 & 0.679 \spm0.047 & 2.016 \spm0.053 & \cellcolor{blue!10}{\textbf{2.205 \spm0.187}} & \cellcolor{blue!10}{\textbf{2.249 \spm0.221}} & 2.229 \spm0.330 & 2.256 \spm0.133 \\
			\midrule
			\multirow{2}{*}{COPD FEV$_1$}
			& DeMMO-var1 & 0.920 \spm0.031 & 0.310 \spm0.042 & 18.870 \spm0.320 & 20.264 \spm0.641 & 19.179 \spm1.203 & 19.778 \spm1.036 & \cellcolor{blue!10}{\textbf{20.579 \spm0.608}} \\
			& DeMMO-var2 & \cellcolor{blue!10}{\textbf{0.914 \spm0.031}} & \cellcolor{blue!10}{\textbf{0.317 \spm0.046}} & \cellcolor{blue!10}{\textbf{18.787 \spm0.331}} & \cellcolor{blue!10}{\textbf{20.158 \spm0.686}} & \cellcolor{blue!10}{\textbf{19.030 \spm1.307}} & \cellcolor{blue!10}{\textbf{19.758 \spm1.025}} & 20.643 \spm0.554 \\
			\bottomrule
		\end{tabular}%
	}
\end{table*}

DeMMO performs best overall. It achieves an nMSE of
$0.769\pm0.021$, compared with $0.773\pm0.035$ for DeMMO-var1 and
$0.772\pm0.033$ for DeMMO-var2, while its wR of $0.464\pm0.023$ exceeds
$0.459\pm0.040$ and $0.459\pm0.038$, respectively. DeMMO also obtains the
best nMSE and wR for both PD outcomes and MS EDSS. DeMMO-var1 is marginally
better for PFF SPPB, whereas DeMMO-var2 is marginally better for COPD
FEV$_1$. At the visit level, DeMMO gives the lowest RMSE in 12 of the 25
tasks, compared with seven for DeMMO-var1 and six for DeMMO-var2.

The two variants retain DeMMO's outcome-specific sparsity, temporal continuity,
and cross-objective relation learning, but use adaptive non-convex reweighting
to reduce the estimation bias induced by the convex Lasso, group-Lasso, and
fused-Lasso penalties. Their lack of a consistent aggregate improvement
suggests that shrinkage-induced bias is not a major limitation in this
setting. We conjecture that this result reflects the substantial
behavioural and measurement noise in free-living DMOs. By weakening shrinkage,
the non-convex penalties may retain more noise together with the DMO signal,
leading to poorer generalisation; the stronger shrinkage of convex DeMMO may
instead provide beneficial regularisation. Convex DeMMO also shows lower
overall variability and requires substantially less computation because it
avoids the additional difference-of-convex iterations. These results favour
the original DeMMO formulation for this noisy longitudinal DMO setting.

\section{Additional Experimental Results}
\label{sec:additional_results}
Tables~\ref{tab:visit_rmse_pd_hy}--\ref{tab:visit_rmse_copd} provide the
numerical results underlying Figure~\ref{fig:rmse_across_visits}. 

\begin{table}[!t]
	\centering
	\caption{Visit-specific RMSE for PD H\&Y (mean $\pm$ standard deviation over five matched participant splits). The best mean in each column is shown in bold.}
	\label{tab:visit_rmse_pd_hy}
		\begin{tabular}{lccccc}
			\toprule
			Method & T1 & T2 & T3 & T4 & T5 \\
			\midrule
			cFSGL & 0.550 \spm0.023 & 0.510 \spm0.042 & 0.557 \spm0.029 & 0.535 \spm0.051 & 0.558 \spm0.037 \\
			FRoTS & 0.543 \spm0.023 & 0.514 \spm0.040 & 0.550 \spm0.027 & 0.531 \spm0.042 & 0.573 \spm0.049 \\
			MAGPP & 0.544 \spm0.021 & 0.522 \spm0.043 & 0.552 \spm0.030 & 0.543 \spm0.040 & 0.586 \spm0.053 \\
			MLP-MSE & 0.556 \spm0.028 & 0.518 \spm0.040 & 0.560 \spm0.032 & 0.542 \spm0.046 & 0.565 \spm0.059 \\
			MLP-L1 & 0.571 \spm0.026 & 0.532 \spm0.034 & 0.572 \spm0.026 & 0.542 \spm0.047 & 0.567 \spm0.036 \\
			Rank-N & 0.577 \spm0.027 & 0.532 \spm0.034 & 0.583 \spm0.033 & 0.539 \spm0.050 & 0.568 \spm0.035 \\
			RankSim & 0.556 \spm0.023 & 0.519 \spm0.038 & 0.558 \spm0.030 & 0.549 \spm0.041 & 0.558 \spm0.047 \\
			ACCon & 0.562 \spm0.019 & 0.528 \spm0.038 & 0.563 \spm0.028 & 0.545 \spm0.052 & 0.569 \spm0.052 \\
			\midrule
			DeMMO & \cellcolor{blue!10}{\textbf{0.538 \spm0.022}} & \cellcolor{blue!10}{\textbf{0.504 \spm0.038}} & \cellcolor{blue!10}{\textbf{0.542 \spm0.025}} & \cellcolor{blue!10}{\textbf{0.524 \spm0.052}} & \cellcolor{blue!10}{\textbf{0.543 \spm0.045}} \\
			\bottomrule
		\end{tabular}%
\end{table}

\begin{table}[!t]
	\centering
	\caption{Visit-specific RMSE for PD MDS--UPDRS (mean $\pm$ standard deviation over five matched participant splits). The best mean in each column is shown in bold.}
	\label{tab:visit_rmse_pd_mds}
		\begin{tabular}{lccccc}
			\toprule
			Method & T1 & T2 & T3 & T4 & T5 \\
			\midrule
			cFSGL & 11.959 \spm0.227 & 11.852 \spm0.077 & 12.299 \spm0.472 & 11.742 \spm0.527 & 11.488 \spm0.227 \\
			FRoTS & 11.946 \spm0.328 & 11.700 \spm0.258 & 12.333 \spm0.500 & 11.674 \spm0.515 & 11.740 \spm0.489 \\
			MAGPP & \cellcolor{blue!10}{\textbf{11.925 \spm0.335}} & 11.901 \spm0.368 & 12.517 \spm0.517 & 11.997 \spm0.485 & 11.722 \spm0.424 \\
			MLP-MSE & 12.161 \spm0.213 & 11.847 \spm0.166 & 12.318 \spm0.636 & 11.865 \spm0.693 & 11.785 \spm0.375 \\
			MLP-L1 & 12.424 \spm0.400 & 12.125 \spm0.426 & 12.256 \spm0.465 & 12.373 \spm0.665 & 12.087 \spm0.330 \\
			Rank-N & 12.213 \spm0.213 & 12.162 \spm0.187 & 12.708 \spm0.931 & 12.165 \spm0.819 & 12.206 \spm0.454 \\
			RankSim & 12.214 \spm0.338 & 11.953 \spm0.172 & 12.503 \spm0.863 & 11.877 \spm0.598 & 11.846 \spm0.371 \\
			ACCon & 12.287 \spm0.258 & 12.009 \spm0.128 & 12.425 \spm0.719 & 12.285 \spm0.686 & 12.260 \spm0.949 \\
			\midrule
			DeMMO & 12.073 \spm0.260 & \cellcolor{blue!10}{\textbf{11.676 \spm0.055}} & \cellcolor{blue!10}{\textbf{12.002 \spm0.489}} & \cellcolor{blue!10}{\textbf{11.428 \spm0.524}} & \cellcolor{blue!10}{\textbf{11.339 \spm0.430}} \\
			\bottomrule
		\end{tabular}%
\end{table}

\begin{table}[!t]
	\centering
	\caption{Visit-specific RMSE for MS EDSS (mean $\pm$ standard deviation over five matched participant splits). The best mean in each column is shown in bold.}
	\label{tab:visit_rmse_ms}
		\begin{tabular}{lccccc}
			\toprule
			Method & T1 & T2 & T3 & T4 & T5 \\
			\midrule
			cFSGL & 0.826 \spm0.053 & 0.845 \spm0.038 & 0.998 \spm0.041 & 1.023 \spm0.028 & 1.108 \spm0.053 \\
			FRoTS & 0.832 \spm0.047 & 0.841 \spm0.043 & 0.997 \spm0.043 & 1.022 \spm0.039 & 1.132 \spm0.061 \\
			MAGPP & 0.831 \spm0.047 & 0.856 \spm0.044 & 1.014 \spm0.045 & 1.035 \spm0.045 & 1.153 \spm0.060 \\
			MLP-MSE & 0.843 \spm0.054 & 0.851 \spm0.053 & 1.018 \spm0.036 & 1.001 \spm0.021 & 1.167 \spm0.075 \\
			MLP-L1 & 0.861 \spm0.052 & 0.869 \spm0.056 & 1.028 \spm0.039 & 1.038 \spm0.052 & 1.137 \spm0.058 \\
			Rank-N & 0.845 \spm0.058 & 0.892 \spm0.046 & 1.050 \spm0.033 & 1.064 \spm0.037 & 1.159 \spm0.051 \\
			RankSim & 0.837 \spm0.054 & 0.854 \spm0.048 & 0.996 \spm0.031 & \cellcolor{blue!10}{\textbf{0.997 \spm0.013}} & 1.165 \spm0.047 \\
			ACCon & 0.844 \spm0.061 & 0.857 \spm0.047 & 1.037 \spm0.034 & 1.020 \spm0.012 & 1.176 \spm0.078 \\
			\midrule
			DeMMO & \cellcolor{blue!10}{\textbf{0.818 \spm0.052}} & \cellcolor{blue!10}{\textbf{0.823 \spm0.039}} & \cellcolor{blue!10}{\textbf{0.984 \spm0.038}} & 1.005 \spm0.032 & \cellcolor{blue!10}{\textbf{1.098 \spm0.062}} \\
			\bottomrule
		\end{tabular}%

\end{table}

\begin{table}[!t]
	\centering
	\caption{Visit-specific RMSE for PFF SPPB (mean $\pm$ standard deviation over five matched participant splits). The best mean in each column is shown in bold.}
	\label{tab:visit_rmse_pff}
		\begin{tabular}{lccccc}
			\toprule
			Method & T1 & T2 & T3 & T4 & T5 \\
			\midrule
			cFSGL & \cellcolor{blue!10}{\textbf{1.961 \spm0.026}} & 2.251 \spm0.119 & 2.293 \spm0.101 & 2.299 \spm0.157 & 2.301 \spm0.106 \\
			FRoTS & 1.967 \spm0.052 & 2.228 \spm0.142 & 2.264 \spm0.075 & 2.489 \spm0.077 & 2.389 \spm0.140 \\
			MAGPP & 1.996 \spm0.051 & 2.225 \spm0.165 & 2.299 \spm0.079 & 2.539 \spm0.077 & 2.433 \spm0.144 \\
			MLP-MSE & 1.979 \spm0.060 & 2.215 \spm0.116 & 2.379 \spm0.091 & 2.484 \spm0.145 & 2.471 \spm0.122 \\
			MLP-L1 & 2.005 \spm0.038 & 2.217 \spm0.099 & 2.379 \spm0.150 & 2.494 \spm0.172 & 2.507 \spm0.132 \\
			Rank-N & 2.028 \spm0.069 & 2.261 \spm0.170 & 2.322 \spm0.098 & 2.479 \spm0.086 & 2.517 \spm0.082 \\
			RankSim & 1.996 \spm0.060 & \cellcolor{blue!10}{\textbf{2.213 \spm0.095}} & 2.332 \spm0.105 & 2.470 \spm0.117 & 2.449 \spm0.092 \\
			ACCon & 2.021 \spm0.059 & 2.224 \spm0.142 & 2.320 \spm0.053 & 2.544 \spm0.115 & 2.553 \spm0.156 \\
			\midrule
			DeMMO & 2.019 \spm0.036 & 2.221 \spm0.121 & \cellcolor{blue!10}{\textbf{2.228 \spm0.109}} & \cellcolor{blue!10}{\textbf{2.236 \spm0.191}} & \cellcolor{blue!10}{\textbf{2.265 \spm0.104}} \\
			\bottomrule
		\end{tabular}%
	
\end{table}

\begin{table}[!t]
	\centering
	\caption{Visit-specific RMSE for COPD FEV$_1$ (mean $\pm$ standard deviation over five matched participant splits). The best mean in each column is shown in bold.}
	\label{tab:visit_rmse_copd}
		\begin{tabular}{lccccc}
			\toprule
			Method & T1 & T2 & T3 & T4 & T5 \\
			\midrule
			cFSGL & 18.927 \spm0.135 & 20.601 \spm0.421 & 19.472 \spm0.809 & 20.009 \spm0.756 & 20.835 \spm0.334 \\
			FRoTS & 19.383 \spm0.274 & 20.828 \spm0.706 & 19.197 \spm0.901 & 20.331 \spm0.884 & 21.238 \spm0.257 \\
			MAGPP & 19.424 \spm0.171 & 20.885 \spm0.720 & 19.337 \spm0.980 & 20.558 \spm1.078 & 21.562 \spm0.176 \\
			MLP-MSE & 19.231 \spm0.739 & 20.899 \spm0.684 & 20.476 \spm0.971 & 20.320 \spm0.735 & 20.851 \spm0.589 \\
			MLP-L1 & 19.241 \spm0.393 & 21.582 \spm0.892 & 20.104 \spm0.986 & 20.795 \spm1.192 & 21.144 \spm0.490 \\
			Rank-N & 19.289 \spm0.527 & 21.263 \spm0.592 & 20.013 \spm0.566 & 20.538 \spm0.715 & 21.078 \spm0.321 \\
			RankSim & 19.187 \spm0.605 & 20.921 \spm0.478 & 19.765 \spm0.723 & 20.192 \spm0.826 & 20.658 \spm0.603 \\
			ACCon & 19.491 \spm0.661 & 21.229 \spm0.738 & 20.004 \spm0.659 & 20.706 \spm0.854 & 20.896 \spm0.518 \\
			\midrule
			DeMMO & \cellcolor{blue!10}{\textbf{18.814 \spm0.152}} & \cellcolor{blue!10}{\textbf{20.237 \spm0.437}} & \cellcolor{blue!10}{\textbf{19.072 \spm0.887}} & \cellcolor{blue!10}{\textbf{19.847 \spm0.654}} & \cellcolor{blue!10}{\textbf{20.478 \spm0.521}} \\
			\bottomrule
		\end{tabular}%
\end{table}

\paragraph{Outcome-specific observations.}
For PD H\&Y, DeMMO is best or tied for best at all five visits, with its clearest
advantages at T2--T4. For PD MDS--UPDRS, it achieves the lowest mean RMSE at
T2--T5, while MAGPP performs best at T1. DeMMO is also consistently
competitive for MS EDSS: it is best at T1--T3 and T5 and remains close to the
best result at T4. The PFF task exhibits a different pattern. cFSGL is
strongest at T1 and RankSim at T2, whereas DeMMO becomes best
from T3 to T5. This later-visit advantage is particularly relevant because the
PFF sample size decreases sharply across follow-up visits, suggesting that
temporal and cross-objective information sharing provides useful
regularisation when outcome-specific data are sparse. For COPD FEV$_1$,
DeMMO achieves the lowest mean RMSE at all five visits, demonstrating that its
advantage also extends to pulmonary-function prediction.

\paragraph{Patterns across visits and methods.}
Across the five outcomes, DeMMO achieves the lowest mean RMSE in 21 of the 25
outcome--visit comparisons. Its performance is not uniformly dominant,
however: other methods lead at PD MDS--UPDRS T1, MS EDSS T4, and PFF T1--T2.
Errors for MS EDSS and PFF
generally increase at later visits, consistent with their declining sample
sizes and potentially greater follow-up heterogeneity, whereas both PD outcomes
show non-monotonic trajectories. Deep regression and ranking methods rarely
lead these visit-specific comparisons and often exhibit larger variability at
later visits. Overall, the tables support selective rather than uniform
information sharing: DeMMO preserves outcome-specific behaviour while gaining
most where longitudinal observations become limited.

\section{Potential Application: Interpretable Feedback for Longitudinal Mobility Monitoring}
\label{sec:clinical_application}
One potential application of DeMMO is an interpretable software system for
longitudinal mobility monitoring. Such software could process repeated
wearable measurements, use DeMMO to estimate a participant's clinical status
at successive visits, and flag a possible decline when the estimated severity
increases or crosses a predefined monitoring boundary. The important question
is then not only \emph{whether} the software detects a change, but also
\emph{how} it explains that change to the user. The following retrospective
analysis illustrates this potential workflow. It does not constitute a
deployed product or a validated clinical alert system.

\paragraph{From DeMMO discovery to software feedback.}
DeMMO assigns a visit-specific coefficient to every DMO and uses stability
selection to identify DMOs that are repeatedly selected across resampled
datasets and hyperparameter settings. For MS EDSS, the two most stable DMOs
are walking-speed P90 and cadence P90 in walking bouts longer than 30 seconds,
with selection probabilities of $0.878$ and $0.874$, respectively. A walking
bout is a continuous period during which a participant is detected as walking
without a substantial interruption. Walking-speed P90 and cadence P90 are the
90th percentiles of walking speed and cadence, respectively, calculated from
such bouts and measured in m/s and steps/min. They reflect relatively high
levels of speed and cadence achieved during sustained real-world walking,
rather than average walking behaviour. Their Pearson and Spearman correlations
are $0.766$ and $0.778$, indicating strongly related but non-identical
information.

In principle, a complete explanation could report the contribution of all 24
DMOs used by DeMMO. However, such an explanation would be lengthy, include
several correlated mobility measures, and place a substantial cognitive burden
on users who may not have computational or clinical expertise. Listing every
contributing variable could therefore obscure the main reason for an alert
rather than make it more understandable. We instead use stability selection
to prioritise a small number of repeatedly identified DMOs and present the two
strongest as a concise user-facing explanation. This is a deliberate trade-off
between completeness and comprehensibility: the full DeMMO model  still use
all 24 DMOs in the background, while the explanation layer summarises the most
stable evidence in a form that a user can readily interpret. The two-DMO
explanation should consequently be understood as a compact summary, not a
claim that the remaining DMOs are irrelevant.

In a potential software pipeline, DeMMO would first indicate that the
participant's estimated status had worsened across visits. The software could
then examine whether these two stable DMOs had also declined or entered ranges
associated with greater impairment, providing a concise explanation for the
alert. We use a deliberately simple, class-balanced decision tree to examine
whether the two DMOs can support such an explanation. The analysis follows the
main experiment's participant-level
$70\%/10\%/20\%$ training/validation/test protocol with seeds 42--46; all
visits from one participant remain in the same split. Maximum depth
($1,2,3,4,$ or $5$) and minimum leaf size ($5,10,20,$ or $30$) are selected by
validation balanced accuracy. The selected tree is then refitted on the
combined training and validation participants and evaluated once on the
held-out test participants.

\paragraph{Binary classification of severe disability.}
We first consider a clinically focused binary task separating severe
disability (EDSS $\geq6.0$) from non-severe disability (EDSS $<6.0$). Higher
EDSS indicates greater impairment, and EDSS 6 marks the stage at which a
walking aid is required for ambulation \citep{kurtzke1983rating}. The
analytical sample contains 2,083 participant--visit observations from 578
participants, comprising 1,157 non-severe and 926 severe observations.

\begin{table*}[!t]
\centering
\caption{Binary MS EDSS classification into non-severe ($<6.0$) and severe
($\geq6.0$) disability using walking-speed P90 and cadence P90. Values are
mean $\pm$ standard deviation over five participant-level test splits.}
\label{tab:two_dmo_edss_binary_classification}
\small
\resizebox{\textwidth}{!}{%
\begin{tabular}{llcccc}
\toprule
Method & Class & Balanced accuracy & Precision & Recall & F1-score \\
\midrule
\multicolumn{6}{l}{\textit{Overall class-balanced performance}} \\
\addlinespace[2pt]
Always most common class & Macro average & 0.500 \spm0.000 & 0.271 \spm0.008 & 0.500 \spm0.000 & 0.351 \spm0.007 \\
Two-DMO decision tree & Macro average & 0.789 \spm0.031 & 0.790 \spm0.034 & 0.789 \spm0.031 & 0.785 \spm0.032 \\
\midrule
\multicolumn{6}{l}{\textit{Class-specific performance}} \\
\addlinespace[2pt]
\multirow{2}{*}{Always most common class}
 & Non-severe & -- & 0.542 \spm0.016 & 1.000 \spm0.000 & 0.703 \spm0.014 \\
 & Severe     & -- & 0.000 \spm0.000 & 0.000 \spm0.000 & 0.000 \spm0.000 \\
\addlinespace[2pt]
\multirow{2}{*}{Two-DMO decision tree}
 & Non-severe & -- & 0.850 \spm0.041 & 0.737 \spm0.047 & 0.788 \spm0.032 \\
 & Severe     & -- & 0.731 \spm0.037 & 0.842 \spm0.057 & 0.782 \spm0.037 \\
\midrule
\multicolumn{6}{l}{\textit{Row-normalised confusion matrix for the two-DMO decision tree}} \\
\addlinespace[2pt]
\multicolumn{4}{l}{True class} & Predicted non-severe & Predicted severe \\
\multicolumn{4}{l}{Non-severe} & 0.737 \spm0.047 & 0.264 \spm0.047 \\
\multicolumn{4}{l}{Severe}     & 0.158 \spm0.057 & 0.842 \spm0.057 \\
\bottomrule
\end{tabular}
}
\end{table*}

The two-DMO tree achieves balanced accuracy of $0.789\pm0.031$ and macro-F1
of $0.785\pm0.032$. It correctly identifies $84.2\%$ of severe observations
and $73.7\%$ of non-severe observations. Among observations predicted as
severe, $73.1\%$ are truly severe. The principal error is that $26.4\%$ of
non-severe observations are flagged as severe, whereas $15.8\%$ of severe
observations are missed. The always-most-common-class baseline, which assigns
every observation to the larger non-severe class without using either DMO,
has balanced accuracy of $0.5$ and cannot identify any severe observations.

\paragraph{Potential feedback after a detected decline.}
The leading divisions of the binary trees are reasonably consistent. Three of
the five fitted trees first divide walking-speed P90 at $0.945$~m/s; the other
two first divide cadence P90 at $97.5$ or $99.5$~steps/min. Within this cohort,
lower values of these sustained-walking measures are more frequently
associated with severe EDSS disability. It is worth noting that these values are descriptive model
splits, not proposed clinical cut-offs.

This result illustrates the type of feedback that the selected DMOs might
support. Rather than returning only an unexplained message such as ``your
mobility may be declining'', a future system could identify the measurements
that contributed to the alert. For example, it might report: ``Your sustained
walking-speed P90 and cadence P90 have decreased since the previous monitoring
period and have moved into ranges that were associated with greater mobility
impairment in the reference data. You may wish to pay closer attention to your
sustained walking performance and discuss a persistent decline with a relevant
professional.'' Such feedback directly connects the software output to
observable changes in daily-life mobility. Importantly, it explains the basis
of the software's judgement rather than the biological cause of the person's
condition.

\paragraph{Extension to three severity classes.}
We next examine whether the same two selected DMOs can address the more
difficult task of distinguishing mild (EDSS $0$--$3.0$), moderate
($3.5$--$5.5$), and severe ($\geq6.0$) disability. The corresponding class
counts are 317, 840, and 926 observations. The model and evaluation protocol
remain unchanged.

\begin{table*}[!t]
\centering
\caption{Three-class MS EDSS classification using walking-speed P90 and
cadence P90. Values are mean $\pm$ standard deviation over five
participant-level test splits; the final panel reports the row-normalised
confusion matrix.}
\label{tab:two_dmo_edss_classification}
\small
\resizebox{\textwidth}{!}{%
\begin{tabular}{llcccc}
\toprule
Method & Class & Balanced accuracy & Precision & Recall & F1-score \\
\midrule
\multicolumn{6}{l}{\textit{Overall class-balanced performance}} \\
\addlinespace[2pt]
Always most common class & Macro average & 0.333 \spm0.000 & 0.153 \spm0.005 & 0.333 \spm0.000 & 0.209 \spm0.005 \\
Two-DMO decision tree & Macro average & 0.593 \spm0.023 & 0.543 \spm0.016 & 0.593 \spm0.023 & 0.533 \spm0.030 \\
\midrule
\multicolumn{6}{l}{\textit{Class-specific performance}} \\
\addlinespace[2pt]
\multirow{3}{*}{Always most common class}
 & Mild     & -- & 0.000 \spm0.000 & 0.000 \spm0.000 & 0.000 \spm0.000 \\
 & Moderate & -- & 0.000 \spm0.000 & 0.000 \spm0.000 & 0.000 \spm0.000 \\
 & Severe   & -- & 0.458 \spm0.016 & 1.000 \spm0.000 & 0.628 \spm0.015 \\
\addlinespace[2pt]
\multirow{3}{*}{Two-DMO decision tree}
 & Mild     & -- & 0.367 \spm0.045 & 0.708 \spm0.082 & 0.481 \spm0.041 \\
 & Moderate & -- & 0.516 \spm0.034 & 0.264 \spm0.075 & 0.344 \spm0.060 \\
 & Severe   & -- & 0.745 \spm0.051 & 0.807 \spm0.054 & 0.774 \spm0.041 \\
\midrule
\multicolumn{6}{l}{\textit{Row-normalised confusion matrix for the two-DMO decision tree}} \\
\addlinespace[2pt]
\multicolumn{3}{l}{True class} & Predicted mild & Predicted moderate & Predicted severe \\
\multicolumn{3}{l}{Mild}     & 0.708 \spm0.082 & 0.230 \spm0.085 & 0.063 \spm0.051 \\
\multicolumn{3}{l}{Moderate} & 0.431 \spm0.095 & 0.264 \spm0.075 & 0.305 \spm0.063 \\
\multicolumn{3}{l}{Severe}   & 0.056 \spm0.043 & 0.137 \spm0.054 & 0.807 \spm0.054 \\
\bottomrule
\end{tabular}
}
\end{table*}

For this harder task, the two-DMO tree achieves balanced accuracy of
$0.593\pm0.023$, macro-precision of $0.543\pm0.016$, and macro-F1 of
$0.533\pm0.030$. It correctly identifies $70.8\%$ of mild and $80.7\%$ of
severe observations, but only $26.4\%$ of moderate observations. The
row-normalised confusion matrix shows that $43.1\%$ of moderate observations
are predicted as mild and $30.5\%$ as severe, whereas direct confusion between
the two extreme classes is less frequent ($6.3\%$ and $5.6\%$). Thus, the two
selected DMOs retain useful information in a finer-grained setting, but they
do not reliably resolve the intermediate disability group.

\paragraph{Scope and limitations.}
These thresholds and classifications are associations learned from Mobilise-D
and should be viewed as hypothesis-generating observations, not diagnostic
cut-offs. The current experiments classify disability at an observed visit;
they do not yet implement the complete software pipeline or prospectively
detect individual deterioration. A practical system would require explicit
rules for defining within-participant change, repeated-measure validation,
personalised reference ranges, and careful design of user-facing feedback. It
would also need to account for factors such as age, body size, walking aids,
comorbidities, and sensor wear conditions. The experiments therefore provide
an initial illustration of how DeMMO-derived patterns could make future
longitudinal mobility-monitoring software more transparent and informative.

\end{document}